\documentclass[manuscript,screen]{acmart}

\AtBeginDocument{%
  \providecommand\BibTeX{{%
    \normalfont B\kern-0.5em{\scshape i\kern-0.25em b}\kern-0.8em\TeX}}}

\usepackage[ruled, lined, linesnumbered, commentsnumbered, longend]{algorithm2e}

\SetCommentSty{mycommfont}
\usepackage{subcaption} 
\usepackage{graphicx}
\newcommand{\tabitem}{~~\llap{\textbullet}~~}

\usepackage{caption}
\usepackage{multirow}
\usepackage{textcomp}
\usepackage{graphicx}

\begin{document}

\title{Word-level Human Interpretable Scoring Mechanism for Novel Text Detection Using Tsetlin Machines}

\author{Bimal Bhattarai}
\email{bimal.bhattarai@uia.no}
\affiliation{%
 \institution{University of Agder}
   \postcode{4879}
 \city{Grimstad}
 \country{Norway}
}

\author{Ole-Christoffer Granmo}
\email{ole.granmo@uia.no}
\affiliation{%
  \institution{University of Agder}
   \postcode{4879}
  \city{Grimstad}
  \country{Norway}}

\author{Lei Jiao}
\email{lei.jiao@uia.no}
\affiliation{%
  \institution{University of Agder}
   \postcode{4879}
  \city{Grimstad}
  \country{Norway}
}

\renewcommand{\shortauthors}{B. Bhattarai, et al.}

\begin{abstract}
    Recent research in novelty detection focuses mainly on document-level classification, employing deep neural networks (DNN). However, the black-box nature of DNNs makes it difficult to extract an exact explanation of why a document is considered novel. In addition, dealing with novelty at the word-level is crucial to provide a more fine-grained analysis than what is available at the document level. In this work, we propose a Tsetlin machine (TM)-based architecture for scoring individual words according to their contribution to novelty. Our approach encodes a description of the novel documents using the linguistic patterns captured by TM clauses. We then adopt this description to measure how much a word contributes to making documents novel. Our experimental results demonstrate how our approach breaks down novelty into interpretable phrases, successfully measuring novelty.
\end{abstract}


\maketitle

\section{Introduction}

The basic principle underlying machine learning classifiers is generalization -- the ability to form a decision boundary that differentiates new input into known classes. When training a supervised classifier, it is common to assume that the classes to be recognized are present both in the training and test data \cite{1}.  However, given an open world, training on all conceivable classes of input is impractical. This problem introduces the need for \emph{novelty detection} -- the task of spotting input classes that one has not seen before.

The problem is particularly severe in text-based supervised classification due to the many-faceted nature of natural language, which gives rise to multiple application-dependent interpretations.  Indeed, researchers have for a long time tried to address novelty detection in natural language. So far, no single best model has appeared. Indeed, the success of each model relies on the properties of each particular dataset.\par

The problem of novelty detection arises in many tasks, such as fault detection \cite{2} and handwritten alphabet recognition \cite{3}. In general, one applies novelty detection when it is required to know whether a given input is similar to the training data or different from it in a significant manner. For natural language text, the novelty detector should discern that a text does not belong to a predefined set of topics. Several challenges make such novelty detection particularly difficult: 
\begin{enumerate}
    \item Textual information tend to be diverse, composed from large vocabularies.
    \item Language and topics are typically evolving, making the novelty detection problem dynamic \cite{4}.
\end{enumerate}
Lately, the above challenges have manifested when using supervised learning for building chatbots, an application area of increasing importance. A chatbot typically needs to handle the language of a multitude of users with evolving information requirements. As such, it must be able to know when it can answer a query and when it faces a new topic. \par

Most of the existing literature on text-based novelty detection addresses one of the following granularity levels:
\begin{enumerate}
    \item Event-level techniques \cite{20} perform topic detection and tracking on a stream of documents.
    \item Document level techniques \cite{22} classify an incoming document as known or novel based on its content.
    \item Sentence-level techniques \cite{21} look for novel sentences within a particular document.
\end{enumerate}

Usually, the sentences/documents are ranked based on some sort of similarity score, obtained from comparing them with previously seen sentences/documents. For instance, the Maximal Marginal Relevance model (MMR) proposed in \cite{23} assigns low scores to previously seen sentences/documents, while novel ones receive high scores. 

\begin{figure}[h]
  \centering
  \includegraphics[width=\linewidth]{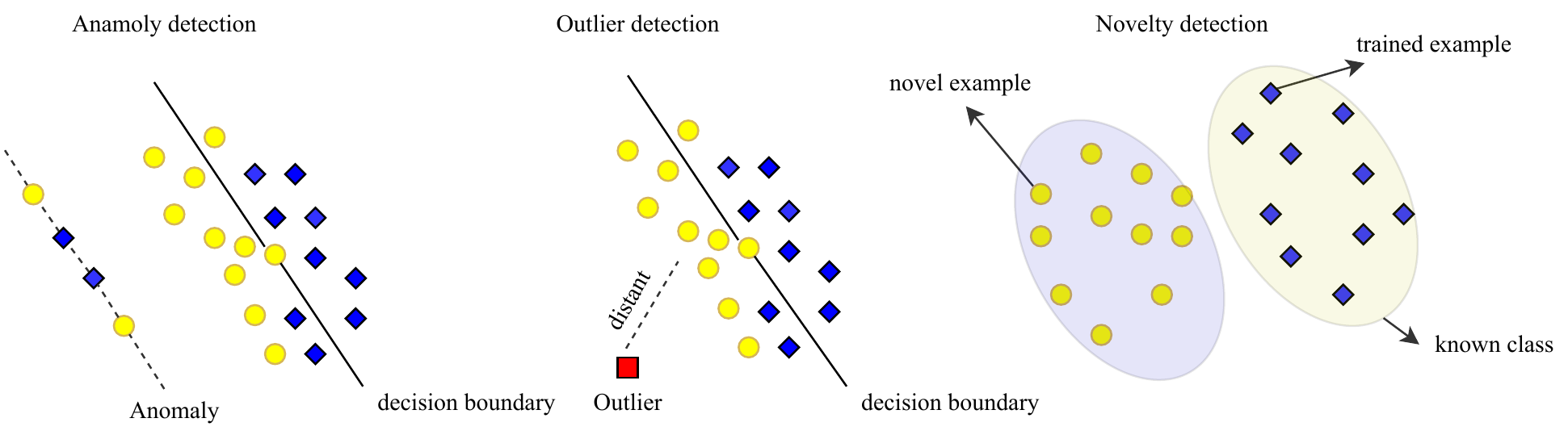}
  \caption{Visualization of outlier detection, anomaly detection and novelty detection.}
  \label{novelty_detection_boundary}%
\end{figure}

Figure \ref{novelty_detection_boundary} illustrates the problem of novelty detection, contrasting it against anomaly- and outlier detection. Anomaly detection concerns discovering anomalies, which are invalid data points. Outlier detection, on the other hand, flags legitimate data points that deviate significantly from the mean. Finally, novelty detection is the discovery of completely new types of data points.

In contrast to previous work, we here focus on novelty detection at the word-level. To this end, we propose a new interpretable machine learning technique for calculating novelty scores for the words within a sentence. The calculation is based on the linguistic patterns captured by a Tsetlin Machine (TM) in the form of AND-rules (i.e., conjunctive clauses). To the best of our knowledge, this is the first study on this problem.

\textbf{Problem Definition}: In the supervised classification setting, $i$ pre-labeled data points $D = \{(x_1, y_1), (x_2,y_2),\ldots,(x_i,y_i)\}$ is used for training. Here, $x_i$ is the $i^{th}$ input example and $y_i$ is its class. The input $x_i$ is an $t$-dimensional real-valued vector $(v_1, v_2, \ldots, v_o) \in \mathbb{R}^t$, where $v_o$ refers to the $o^{th}$ element of the vector. The class $y_i \in Y = \{1, 2, \ldots, C\}$, in turn, is an integer class index referring to one out of $C$ classes.  Learning a classifier means forming a classification function $f(x;D)$, $f: \mathbb{R}^t \rightarrow Y$, based on the data $D$. The function simply assigns a label $y$ to the data point $x$. Our focus is novelty scoring, which can be seen as another function $z(x;D), z: \mathbb{R}^t \rightarrow \mathbb{R}$. The function  calculates a real-valued novelty score for input data point $x$, with the purpose of discerning new classes not found in $Y$. In this way, a classifier can return the correct class label while flagging novel examples. Considering each element in $x$ to represent a specific word, this paper further introduce a method for breaking down the overall score $z(x;D)$ for $x$ into the contribution of each element $v_o$. By doing so, we break down novelty into interpretable phrases.\par

\textbf{Paper contributions:}
In this paper, we use the TM to form conjunctive clauses in propositional logic. In this manner, we capture frequent patterns in the data $D$, which we use to comprehensively characterize the known classes $Y$. The novelty score is then calculated based on examining the clauses that match the given input. By further looking into the composition of each clause, we are able to break down the novelty score into the contribution of the different phrases. This decomposition is based on training clauses for the novel data and then measuring the relative frequency of each word inside the clauses for the known classes, contrasted against the relative frequency obtained from the clauses of the novel input. These scores can in turn be adopted as input features to machine learning classifiers for novelty detection. Similarly, contextual scores can be calculated simply by inspecting the clauses that each word appear in, getting a local view for both novel and known classes.

The remainder of the paper is organized as follows. In Section~\ref{related}, we first summarize related work before we present the details of the TM in Section~\ref{TMA}. This forms the basis for our novelty description architecture, covered in Section~\ref{novelty_description}. In Section~\ref{results}, we present our empirical results, concluding the work in the last section. 

\section{Related work}\label{related}
Several studies have been carried out on supervised multiclass classification in a closed-world setting~\cite{5}. Work addressing open-world settings is more sparse~\cite{6}, with distance-based methods being one of the earliest approaches~\cite{7}. These methods use nearest neighbor search, which leads to scalability problems for larger datasets. Another class of methods are based on single-class classifiers. These includes One-Class SVM \cite{8} and SVDD \cite{9}. Further,  the decision score from SVM has been used to produce a probability distribution for novelty detection~\cite{10}. As no negative training samples are used, single-class classifiers struggle with maximizing the class margin. To overcome the problem of One-Class SVMs, a new learning method named center-based similarity space (CBS) was proposed  in~\cite{11}, which transforms each document in a closed boundary to a central similarity vector that can be used in a binary classifier. \par

Probabilistic methods have also been utilized for novelty detection~\cite{12}. In~\cite{13}, a technique to threshold the entropy of the estimated class probability distribution is proposed. In that method, choosing the entropy threshold needs prior knowledge. Further, the class probability distribution can be misleading when novel data points fall far from the decision boundary. In \cite{14} and \cite{15}, an active learning model is proposed to both discover and classify novel classes during training. However, the appearance of novel instances during testing is not considered. \par

 Recently, DNNs have been used to address the problem of novelty detection.   In~\cite{16}, a two-class SVM classifier is adopted to categorize known and novel classes.  An adversarial sample generation (ASG) framework \cite{17} is used to generate positive and negative samples. Similarly, \cite{18} employs generative adversarial networks (GANs), where the generator produces a mixture of known and novel data. The generator is trained with so-called feature matching loss, and the discriminator performs simultaneous classification and novelty detection. In computer vision, the problem of novel image detection is addressed by introducing the concept of open space risk \cite{1}. This is achieved by reducing the half-space of a binary SVM classifier with two parallel hyperplanes that bound the positive region. Although the positive region is reduced to half-spaces by the binary SVM, their open space risk is still infinite. In \cite{5}, a method called OpenMAX is proposed, which estimates the probability of an input belonging to a novel class. In general, the major weaknesses of these methods are high computational complexity and uninterpretable inference.

\section{Tsetlin Machine (TM) Architecture}\label{TMA}

The TM, proposed in \cite{granmo2018}, is a recent approach to pattern classification, regression, and novelty detection \cite{granmo2019convtsetlin,abeyrathna2019,bhattarai2021novelty}. It captures the frequent patterns of the learning problem using conjunctive clauses in propositional logic. Each clause is a conjunction of literals, where a literal is a propositional/Boolean variable or its negation. Recent research reports that the TM performs competitively with state-of-the-art deep learning networks in text classification~\cite{berge2019, yadav2021sentiment,yadav2021dwr,saha2020causal}. Further, theoretical studies have uncovered robust convergence properties~\cite{zhang2020convergence, jiao2021convergence}.

A basic TM takes a vector $X= (x_1, \ldots, x_o) \in \{0,1\}^o$ of $o$ Boolean features as input. For text input, it is typical to  booleanize the text to form a Boolean set of words, as suggested in \cite{berge2019}. The input features along with their negated counterparts, $\bar{x} = \lnot x = 1-x$ , form a literal set $L = \{x_1, \ldots, x_o, \neg x_1, \ldots, \neg x_o\}$. For classification problems, the sub-patterns associated with the classes are captured by the TM using $m$ conjunctive clauses $C_j^{+}$ or $C_j^{-}$. The $j= 1,\ldots, m/2$ subscript denotes the clause index, while the superscript flags the \emph{polarity} of a clause. In brief, half of the clauses are assigned positive polarity, i.e., $C_j^{+}$, and the other half are assigned negative polarity, i.e., $C_j^{-}$. The positive polarity clauses vote for the input belonging to the class favored by the TM, while the negative polarity clauses vote against that class, that is, for other classes.

\begin{figure}[h]
  \centering
  \includegraphics[width=0.75\linewidth]{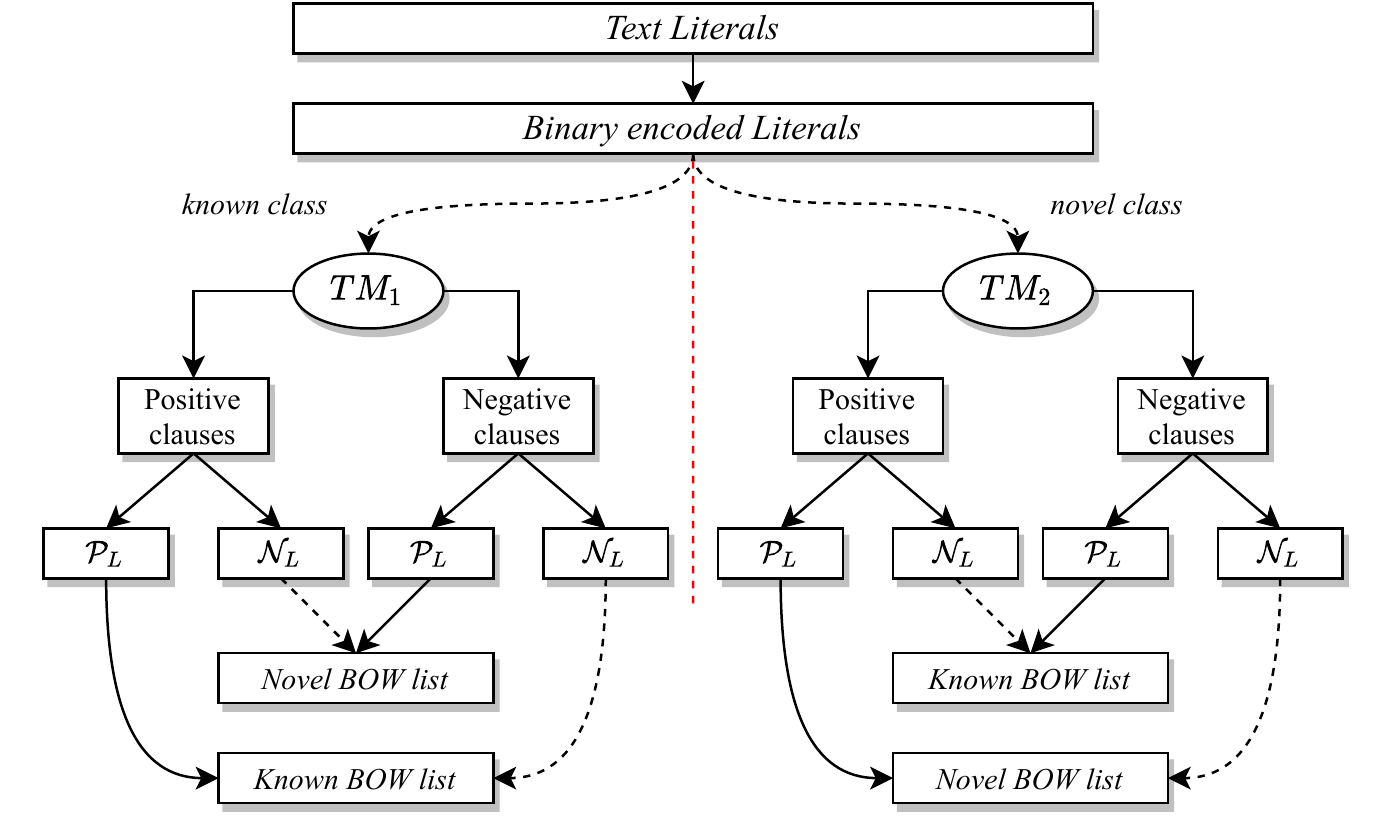}
  \caption{Tsetlin Machine architecture for generating word sequences.}
  \label{novelty_description_architecture}%
\end{figure}

A clause $C_j^\xi, \xi \in \{-,+\},$ is formed by ANDing a subset $L_j^\xi \subseteq L$ of the literal set. That is, the set of literals for clause $C_j^\xi$ with polarity $\xi$ can be written as:
\begin{equation}
    C_j^\xi (X)=\bigwedge_{l \in L_j^\xi} l = \prod_{l \in L_j^\xi} l.
\end{equation}
The clause evaluates to $1$ if and only if all of the literals of the clause also evaluate to $1$. For example, the clause $C_j^\xi(X) = x_1 x_2$ consists of the literals $L_j^\xi = \{x_1, x_2\}$ and outputs $1$, if $x_1 = x_2 = 1$. The final classification decision is obtained by subtracting the negative votes from the positive votes, and then  thresholding the resulting sum using the unit step function $u$: 
\begin{equation}
    \hat{y} = u\left(\sum_{j=1}^{m/2} C_j^+(X) - \sum_{j=1}^{m/2} C_j^-(X)\right).
\end{equation}
For example, the classifier $\hat{y} = u\left(x_1 \bar{x}_2 + \bar{x}_1 x_2 - x_1 x_2 - \bar{x}_1 \bar{x}_2\right)$ captures the XOR-relation. 

For learning, the TM employs a team of Tsetlin Automata (TA), one TA per literal $l \in L$. Each TA performs one of two actions: either \emph{include} or \emph{exclude} its designated literal. The decision whether to include the literal is based on reinforcement: Type I feedback is designed to
produce frequent patterns, while Type II feedback increases the discriminating power of the patterns (see ~\cite{granmo2019convtsetlin} for details). The feedback guides the complete system of TAs towards a Nash equilibrium. At any point in the training process, we have $m$ conjunctive clauses per class, half of them positive and half of them negative. After training is completed, these can be extracted and deployed.\par

\section{Novelty description} \label{novelty_description}
By novelty description, we mean the task of characterizing novel textual content at the word level. For instance, the known content may be mobile phone reviews, and the novel content may be grocery store reviews. For this example, one can characterize the novel content by those words related to grocery stores. However, describing novelty at the word level is nontrivial because the meaning of words varies depending on the context they appear in. For example, let us consider the word ``apple''. This word typically manifests in two different contexts --- it can denote either ``fruit'' or ``cell phone''. Likewise, the word ``bank'' can refer to ``riverbank'' or ``cash bank''. That is, when we consider contextual meaning, the novelty of the word ``apple'' and ``bank'' can be different based on their respective uses. Hence, measuring and describing novel content is a challenging problem.\par

In general, one can detect and characterize novel content by contrasting against the probability of observing textual content $X$, given that the content is known. We denote this probability distribution by $p_{\mathit{known}}(X)$. Assume that the corresponding probability distribution $p_{\mathit{novel}}(X)$ for novel content also is available.  Then, the optimal novelty detection test for a given false positive rate ($\alpha$) can be obtained by thresholding the likelihood ratio $p_{\mathit{novel}}(X)/p_{\mathit{known}}(X)$ \cite{24}. 

Since neither $p_{\mathit{known}}(X)$ or $p_{\mathit{novel}}(X)$ are available to us, we need to estimate them from training examples. Inspired by the work in~\cite{25} on Semi-Supervised Novelty Detection (SSND), we use two sets of examples. One set represents known content and one set represents novel content. We obtain these sets by employing a binary classifier that can distinguish between known and novel content, such as the one we proposed in \cite{bhattarai2021novelty}.

\subsection{Identifying Novel Word Candidates}

In our approach, we first train a TM on input texts represented as Boolean bag-of-words, i.e., as word \emph{sets}. A propositional variable represents each word in the vocabulary, capturing the presence/absence of the corresponding word in the input text. We group the texts into two classes, \emph{Known} and \emph{Novel}. The first represents known content, and the second represents novel content. Our task is to describe how the second group of text is novel at the word level. To this end, we first identify novel word candidates, followed by scoring and ranking the words based on their contribution to novelty.

Figure \ref{novelty_description_architecture} shows our architecture for identifying novel word candidates. As seen, after training, we obtain the clauses of the two classes, \emph{Known} and \emph{Novel}. For each class, we extract all those words that the class' clauses include. Each clause contains a combination of both plain ($\mathcal{P}_L$) and negated ($\mathcal{N}_L$) words. As such, the plain and the negated words serve two different roles. The plain words characterize the corresponding class, while the negated words characterize the other class. We exploit this property as follows, building two bag-of-words (BOW). The first is a bag of known words, referred to as $\mathcal{B_K}$, and the second is a bag of novel words, referred to $\mathcal{B_N}$.

For class \emph{Known}, we perform the following procedure:
\begin{itemize}
    \item We consider the words included in positive clauses first. Here, the plain words $\mathcal{P}_L$ are added to the bag $\mathcal{B_K}$ of known words, while the negated words are placed in the bag of novel words $\mathcal{B_N}$.
    \item For negative clauses we do the opposite. The plain words $\mathcal{P}_L$  are added to the novel words bag $\mathcal{B_N}$. The negated words $\mathcal{N}_L$, on the other hand, are added to the known word bag $\mathcal{B_K}$.
\end{itemize}
The above procedure is inverted for class \emph{Novel}:
 \begin{itemize}
     \item For the positive clauses, the plain words $\mathcal{P}_L$ are added to the novel word bag $\mathcal{B_N}$, while the negated words are added to the known word bag $\mathcal{B_K}$.
     \item Conversely, for the negative clauses, the plain words are added to $\mathcal{B_K}$, characterizing the known class, while the negated words $\mathcal{N}_L$ are added to $\mathcal{B_N}$.
 \end{itemize}

\subsection{Scoring Word Novelty}

With the word bags $\mathcal{B_K}$ and $\mathcal{B_N}$ available, we calculate novelty scores at the word level as follows. From the unique words in the bags $\mathcal{B_K}$ and $\mathcal{B_N}$, we produce two corresponding word sets, $\mathcal{S_K}$ and $\mathcal{S_N}$. Assume these respectively contain $K$ and $N$ unique words:
\begin{equation}
    \begin{split}
        \mathcal{S_K} = \{s_1, s_2,\ldots, s_k,\ldots, s_K\}, \\
        \mathcal{S_N} = \{s_1, s_2, \ldots, s_n,\ldots, s_N\}.
    \end{split} \label{equation1}
\end{equation}
Here, $s_k$ represents a specific word in the set $\mathcal{S_K}$, while $s_n$ represents a specific word in the set $\mathcal{S_N}$.

We next estimate the occurrence probability $p_{s_i}$ of each word $s_i$ in $\mathcal{S_K}$, from the known class. The estimate is based on the relative frequency of $s_i$ in the word bag $\mathcal{B_K}$ as given by Eq. (\ref{eqn:relative_frequency}):
\begin{equation}
    p_{s_i}^{\mathcal{K}} = \frac{\mathcal{F}_i^{\mathcal{K}}}{\sum_{k=1}^K \mathcal{F}_k^{\mathcal{K}}}.
    \label{eqn:relative_frequency}
\end{equation}
Here, $\mathcal{F}_i^{\mathcal{K}}$ is the frequency of word $s_i$ in  $\mathcal{B_K}$, i.e., the number of times that word $s_i$ has the appropriate role in one of the clauses (as defined in the previous section). To prevent infinite or zero scores, we assume that every word has a minimum frequency of $1$. In the following, we denote the set of relative frequencies for the words from  $\mathcal{B_K}$ by $p_\mathcal{K}$, while $p_\mathcal{N}$ is the set of relative frequencies for the words from $\mathcal{B_N}$, as captured by Eq.~(\ref{equation3}):
\begin{equation}
\begin{split}
    p_\mathcal{K}= \{p_{s_1}^{\mathcal{K}}, p_{s_2}^{\mathcal{K}}, \ldots, P_{s_K}^{\mathcal{K}}\},\\
    p_\mathcal{N}= \{p_{s_1}^{\mathcal{N}}, p_{s_2}^{\mathcal{N}},\ldots, p_{s_N}^{\mathcal{N}}\}.
\end{split} 
\label{equation3}
\end{equation}

\begin{figure}[h]
  \centering
  \includegraphics[width=0.8\linewidth]{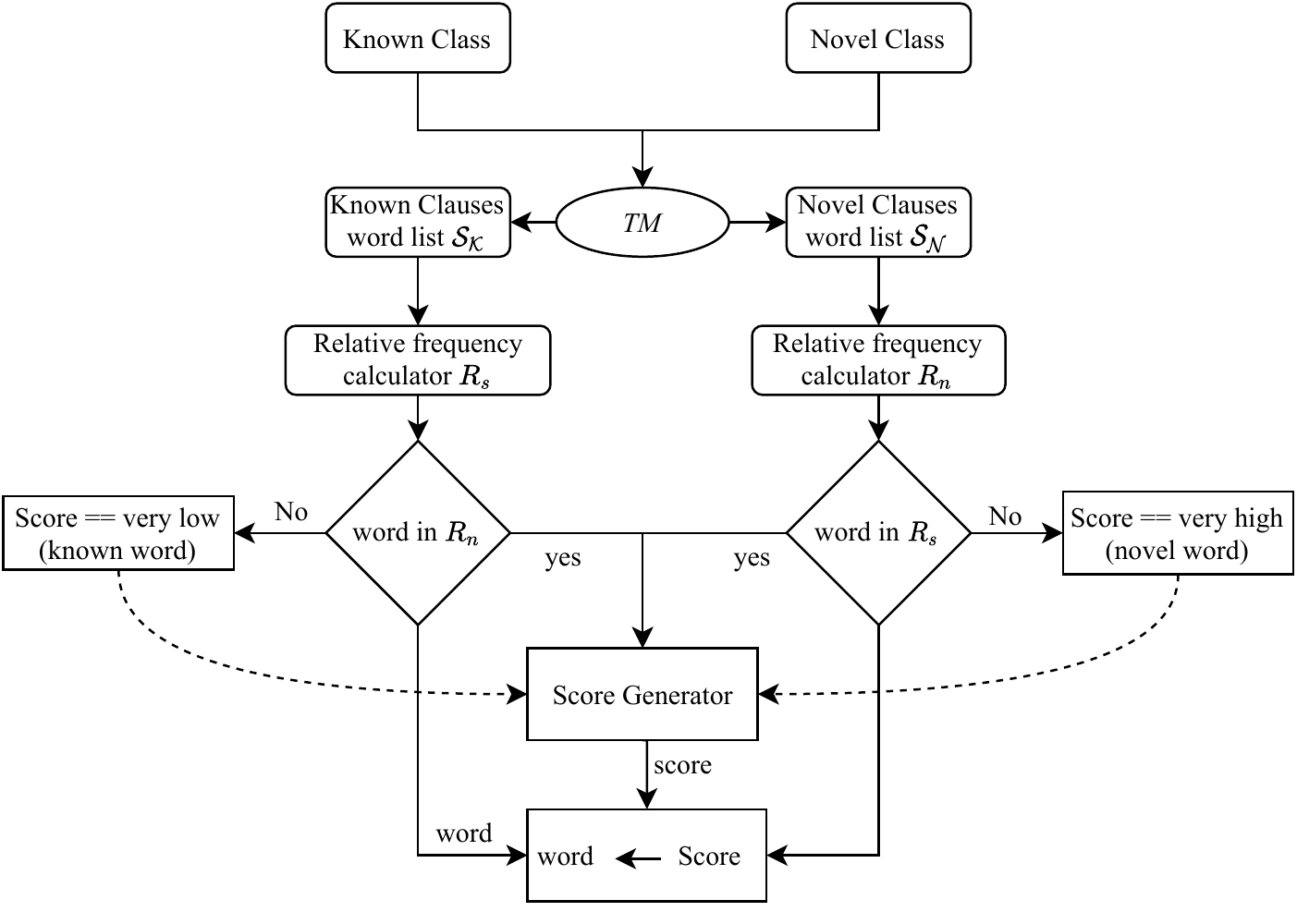}
  \caption{Novelty scoring calculation for each word.}
  \label{scoring_Architecture}%
\end{figure}

The calculation of the novelty score for each word  depends on whether $s_k \in \mathcal{S_K}$, $s_k \in \mathcal{S_N}$, or both, as shown in Eq.~(\ref{equation_novelty_score}):
\begin{equation}
    \mathit{Score}(s_i) =\begin{cases}
    \frac{p_{s_i}^{\mathcal{N}}}{p_{s_i}^{\mathcal{K}}}&\mbox{if } s_k \in \mathcal{S_K} \cap \mathcal{S_N},\\
    0&\mbox{if } s_k \in \mathcal{S_K} \setminus \mathcal{S_N},\\
    \infty&\mbox{if } s_k \in \mathcal{S_N} \setminus \mathcal{S_K}.
    \end{cases}\label{equation_novelty_score}
\end{equation}
Here, $p_{s_i}^{\mathcal{N}}$ and $p_{s_i}^{\mathcal{K}}$ denote the estimated occurrence probabilities of the word $s_i$ from $p_\mathcal{N}$ and $p_\mathcal{K}$, respectively.
The score defines how much a word contributes in a sentence/document to make it novel. That is, a higher score signals higher novelty and vice versa. Figure~\ref{scoring_Architecture} shows the resulting TM-based architecture and flow of information for the above scoring approach. \par

To capture multiple word meanings, decided by context, we also propose a contextual scoring approach. We assume that words that appear in the same clause are related semantically, and accordingly, we use clause co-occurrence of words to measure semantic relations. The intent is to be able to differ between, for example, the meaning of ``apple'' in ``apple phone'' and the meaning of ``apple'' in  ``apple fruit''. We achieve this through leveraging clauses that capture ``apple'' and ``phone'' in combination with other clauses that capture ``apple'' and ``fruit''.

The scoring is again performed in two steps:
\begin{enumerate}
    \item Rather than measuring the frequency of individual words, we now measure frequency of co-occurrence among the TM clauses. For instance, let us consider the word pair $(s_1, s_2)$ and novel class, associated with a total number of $m$ clauses. The frequency of the word pair occurring together in the clauses is then given as:
    \begin{equation}
        p_{s_1, s_2}^\mathcal{N}=\frac{\mathcal{F}_{s_1,s_2}^\mathcal{N}}{m}.
    \end{equation}
    Here, $F_{s_1, s_2}^\mathcal{N}$ is the number of times the word pair occur together across the $m$ clauses of the novel class.
    \item Finally, the contextual score for the word pair ($s_1, s_2$) in class \emph{Novel} can be defined as:
    \begin{equation}
        \mathit{Score}_{\mathit{context}}^\mathcal{N}(s_1, s_2) = \frac{p_{s_1, s_2}^\mathcal{N}}{p_{s_1}^\mathcal{N} \times p_{s_2}^\mathcal{N}}.
    \end{equation}
    Above, $p_{s_2}^\mathcal{N}$ and $p_{s_1}^\mathcal{N}$ are the individual frequencies of each word across the novel clauses, from the previous subsection.
 \end{enumerate}
Notice how the above score increases with lower individual frequencies as well as with higher joint frequency, measuring dependence over the clauses. In the same way, we can calculate dependence over the clauses for the known class as well.

\subsection{Case Study}
We now demonstrate our novelty description approach, steb-by-step, using two example sentences from the sports domain. For illustration purposes, we consider the class Cricket to be \emph{Known} and the class Rugby to be \emph{Novel}.
\begin{itemize}
\item \textbf{Class} : Cricket (Known)\\
\textbf{Text}: England won the cricket match by hitting six in the last ball.\\
\textbf{Words}: \textit{``England''}, \textit{``won''}, \textit{``cricket''}, \textit{``match''}, \textit{``hit''}, \textit{``six''}, \textit{``ball''}.
\item \textbf{Class}: Rugby (Novel)\\
\textbf{Text}: England won the rugby match despite using old ball.\\
\textbf{Words}: \textit{``England ''}, \textit{``won''}, \textit{``rugby''}, \textit{``match''}, \textit{``despite''}, \textit{``old''}, \textit{``ball''}.
\end{itemize}
We first create the set of $10$ unique  words $W = \{\mathit{``England''}, \mathit{``won''}, \mathit{``cricket''}, \mathit{``match''}, \mathit{``hit''}, \mathit{``six''}, \mathit{``ball''}, \mathit{``rugby''}, \mathit{``despite''},$ $\mathit{``old''}\}$ from the words in the two sentences, each with a unique index $o$. From this set, we produce the input feature vector for the TM, $X= [x_1, x_2, \ldots, x_{10}]$. Each propositional input $x_o$ in $X$ refers to a particular word. Jointly, the propositional inputs are used to represent an input text. If a word $w_o \in W$ is present in the document, the corresponding propositional input $x_o$ is set to $1$, otherwise, it is set to $0$.

After TM training, we obtain a set of clauses,  as examplified in Table \ref{caluses_visualization}. The clauses $(C_1^+)_\mathcal{K}$, $(C_2^+)_\mathcal{K}$, $(C_1^-)_\mathcal{N}$, $(C_2^-)_\mathcal{N}$ vote for class \emph{Known}, while $(C_1^-)_\mathcal{K},~(C_2^-)_\mathcal{K},~(C_1^+)_\mathcal{N},~ (C_2^+)_\mathcal{N}$ vote for class \emph{Novel}. These clauses are then used to produce two bag-of-words, $\mathcal{B^K}$ and $\mathcal{B^N}$. All the plain words in $(C_1^+)_\mathcal{K}$, $(C_2^+)_\mathcal{K}$, $(C_1^-)_\mathcal{N}$, $(C_2^-)_\mathcal{N}$ are placed in $\mathcal{B^K}$, while all the negated words are placed in $\mathcal{B^N}$. Since none of the words are negated in the clauses, we now have $\mathcal{B^K} = (\mathit{``England''}, \mathit{``cricket''}, \mathit{``match''}, \mathit{``hit''}, \mathit{``six''}, \mathit{``cricket''}, \mathit{``six''},$ $\mathit{``cricket''}, \mathit{``won''},$ $\mathit{``six''}, \mathit{``ball''}, \mathit{``cricket''}, \mathit{``hit''}, \mathit{``six''})$. Correspondingly, all the plain words in $(C_1^-)_\mathcal{K},~(C_2^-)_\mathcal{K},~(C_1^+)_\mathcal{N},~ (C_2^+)_\mathcal{N}$ are placed in $\mathcal{B^N}$, while all the negated words are placed in $\mathcal{B^K}$.

Within each bag-of-words, each word occurs with a certain frequency. For instance, the word \textit{``match''} occurs once in $\mathcal{B^K}$ and twice in $\mathcal{B^N}$. Notice that the total number of word occurrences are different for each class -- $14$ words in class \emph{Known} and $13$ words in class \emph{Novel}. Hence, the relative frequency for  \textit{``match''} in class \emph{Known} becomes $p^\mathcal{K}_\mathit{match} = \frac{1}{14} = 0.071$ while for class \emph{Novel} it becomes  $p^\mathcal{N}_\mathit{match} = \frac{2}{13} = 0.154$. Table~\ref{frequency_probability} lists the frequencies of the words per class.

\begin{table}
    \centering
    \caption{Clauses with conjunctive word patterns for known and novel class}
    \label{caluses_visualization}
    \begin{tabular}{c|c}
    \toprule
    Known Clauses & Novel Clauses \\
    \midrule
    \tabitem $(C_1^+)_\mathcal{K} = \textit{``England''} \wedge \textit{``cricket''}\wedge \textit{``match''}\wedge \textit{``hit''}\wedge\textit{``six''}$ & \tabitem $(C_1^+)_\mathcal{N} = \textit{``England''} \wedge \textit{``won''}\wedge \textit{``rugby''}\wedge \textit{``old''}$ \\
    \tabitem $(C_1^-)_\mathcal{K} = \textit{``won''}\wedge \textit{``rugby''}\wedge \textit{``ball''}$ & \tabitem $(C_1^-)_\mathcal{N} = \textit{``cricket''} \wedge \textit{``won''}\wedge \textit{``six''}\wedge \textit{``ball''}$\\
    \tabitem $(C_2^+)_\mathcal{K} = \textit{``cricket''} \wedge \textit{``six''}$ & \tabitem $(C_2^+)_\mathcal{N} = \textit{``rugby''} \wedge \textit{``match''}\wedge \textit{``despite''}\wedge \textit{``old''}$\\
    \tabitem $(C_2^-)_\mathcal{K} = \textit{``rugby''} \wedge \textit{``match''}$ & \tabitem $(C_2^-)_\mathcal{N} = \textit{``cricket''} \wedge \textit{``hit''}\wedge\textit{``six''}$ \\
    \bottomrule
\end{tabular}
\end{table}

We are now ready to calculate the novelty score for each word in $W$. Let us consider the word ``rugby'' from the novel word set and the word ``cricket'' from the known word set. For ``rugby'', we first calculate its relative frequency (\ref{eqn:relative_frequency}). In the bag-of-word $\mathcal{B_N}$ for class \emph{Novel}, ``rugby'' occurs \textit{four} times, i.e., $\mathcal{F}^\mathcal{N}_{\mathit{rugby}} = 4$. Since we assume that a word has a minimum frequency of $1$, we further have 
$\mathcal{F}^\mathcal{K}_{\mathit{rugby}} = 1$, despite ``rugby'' not appearing in the text from class \emph{Known}.

From Table~\ref{user_case_table}, we observe that the total word frequencies for the known and novel classes are $14$ and $13$, respectively. Hence, the relative frequencies for ``rugby'' becomes $p_{rugby}(\mathcal{K}) = 0.307$ for class \emph{Known} and  $p_{rugby}(\mathcal{N})= 0.071$ for class \emph{Novel} (Eqn. \ref{eqn:relative_frequency}).

Because the clauses characterize each class \emph{Known} and \emph{Novel}, notice how ``rugby'' gets the relatively high novelty score $Score_{rugby} = 4.651$. That is, its relative frequency is high in the novel class and low in the known class. Conversely, the word ``cricket'' is repeated \textit{four} times in $\mathcal{B^K}$ and \textit{once} in $\mathcal{B^N}$. Its relative frequencies thus becomes $p_{cricket}(\mathcal{K}) = 0.28$ for class \emph{Known} and $p_{cricket}(\mathcal{N})= 0.076$ for class \emph{Novel}. Accordingly, the novelty score becomes $Score_{cricket} = 0.271$, which is a low score denoting a strong inclination of the word towards the known class.

Overall, Table~\ref{user_case_table} shows how the words characterizing class \emph{Known} get a relatively low novelty score, while those characterizing class \emph{Novel} obtain high scores.

\begin{table}
    \caption{Relative frequency and score for each word}
    \label{frequency_probability}
    \begin{tabular}{cccc|cccc}
    \toprule
    \multicolumn{4}{c}{Known } & \multicolumn{4}{c}{Novel}\\
    Word & Frequency & Relative frequency & Score & Word & Frequency & Relative frequency & Score\\
    \midrule
    England & 1 & 0.071 & 1.070 & England & 1 & 0.076 & 1.070\\
    Won & 1 & 0.071 & 2.169 & Won & 2 & 0.154 & 2.169\\
    Cricket & 4 & 0.28 & 0.271 & Rugby & 4 & 0.307 & 4.651\\
    Match & 1 & 0.071 & 2.169 & Match & 2 & 0.154 & 2.169\\
    Hit & 2 & 0.142 & 0.535 & Despite & 1 & 0.076 & 1.15\\
    Six & 4 & 0.28 & 0.271 & Old & 2 & 0.153 & 2.31\\
    Ball & 1 & 0.071 & 1.070 & Ball & 1 & 0.076 & 1.070\\
  \bottomrule
\label{user_case_table}
\end{tabular}
\end{table}

\section{Results and Discussion}\label{results}
In this section, we evaluate our proposed novelty description approach on two publicly available datasets: \emph{BBC Sports} and \emph{Twenty Newsgroups}. We further explore how effective our model is at producing discriminative novelty scores at the word level using TM clauses. 

\subsection{Baseline}
A commonly used method to analyze the importance of a word is term frequency-inverse document frequency (TF-IDF) \cite{31}. TF-IDF weighs each word to statistically measure the significance of the word in a given document. To this end, TF-IDF consists of two factors: normalized term frequency (TF) and inverse document frequency (IDF). TF measures the frequency of the word in the document, whereas IDF measures the uniqueness of the word across documents:
\begin{equation}
    TF-IDF_s = \frac{\mathcal{F}_s}{\mathcal{F}} \times \log_2 \frac{|D|}{|D_s| + 1}.
\end{equation}
Here, $\mathcal{F}_s$ is the frequency of the word $s$ in the target document, $\mathcal{F}$ is the sum of the target document word frequencies, $|D|$ is the total number of documents, and $|D_s|$ is the number of  documents containing the word $s$.

In the following, we compare the scoring mechanism of our framework with TF-IDF as a baseline.  To make the comparison as fair as possible, we calculate TF separately for the known and novel classes. IDF, on the other hand, is calculated using all of the documents from both classes (to suppress common words such as stop words). Unlike TF-IDF, even if a word is present in most of the documents, our scoring considers both relevance and context. For example, if a word from class \emph{Novel} also is present in class \emph{Known}, our model is still able to give more weight to that word. This happens when the word, while \emph{syntactically} the same in both classes, have a novel meaning in the novel class, appearing in a novel context. The latter contextual information is captured through those clauses of the novel class that  trigger for that word. TF-IDF is not context-aware, as such.

For comparison, we plot the cumulative frequency distribution (CFD) for the scores of (i) the words only found in the novel dataset, (ii) the words only found in the known dataset, and (iii) the words shared by both datasets. In brief, the CFD shows that the word scores produced by TF-IDF for both known and novel classes are very similar. Thus, TF-IDF does not provide enough discrimination power to distinguish between the two types of words.

\subsection{BBC sports dataset}
The BBC sports dataset contains $737$ documents from the BBC sport website organized in five sports article categories, collected from 2004 to 2005. The resulting vocabulary encompasses $4~613$ terms.  For our experiment, we consider the classes ``cricket'' and ``football'' to be known and the class ``rugby'' to be novel, thus creating an unbalanced dataset. For preprocessing, we perform tokenization, stopword removal, and lemmatization. We run the TM for 100 epochs with $10~000$ clauses, a voting margin $T$ of 50, and a sensitivity $s$ of 25.0.

We present overall novelty score statistics for the words captured by the clauses in Table \ref{bbc_table}. The table shows that class \emph{Novel} words have distinctively higher scores on average than the words from class \emph{Known}. Also notice that the shared words have the highest mean and standard deviation. As analyzed further below, this is the case because the TM will particularly use those words when forming the decision boundary between the two classes. As a result, the shared words will be present in more clauses as characterizing class features. That is the clauses will either single out the words in one class or suppress the words in the other class.

\begin{table}
  \caption{Overall word statistics  for BBC sport dataset}
  \label{bbc_table}
  \begin{tabular}{cccc}
    \toprule
    Category & Total word count & Average score & Standard deviation \\
    \midrule
    Known words & 6660 & 0.74 & 0.23 \\
    Novel words & 1941 & 1.3125 & 3.75\\
    Shared words & 3135 & 11.30 & 316.93\\
  \bottomrule
\end{tabular}
\end{table}

\begin{table}
  \caption{Composition of shared words in BBC Sport}
  \label{overlapped_bbc}
  \begin{tabular}{cccc}
    \toprule
    Composition & Total word count & Average score & Standard deviation \\
    \midrule
    Known words & 10 & 0.11 & 0.070 \\
    Novel words & 17 & 1941.13 & 3919.02\\
    Common words & 3051 & 1.03 & 0.99\\
  \bottomrule
\end{tabular}
\end{table}

\begin{figure}[h]
    \centering
    \begin{subfigure}[b]{0.4\textwidth}
        \includegraphics[width=1\linewidth]{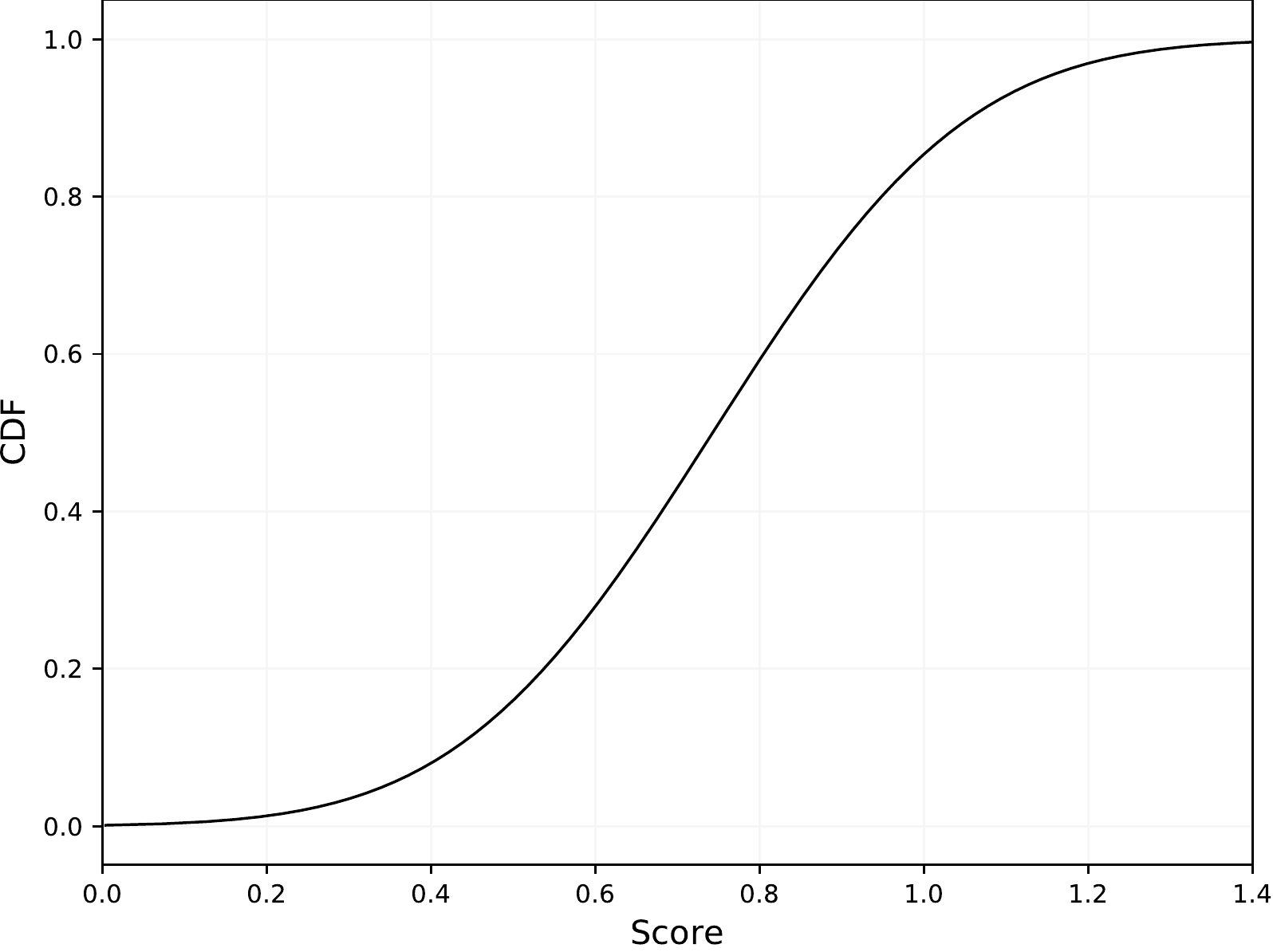}
        \caption{for known words}
        \label{known_cdf}
    \end{subfigure}
    \begin{subfigure}[b]{0.4\textwidth}
        \includegraphics[width=1\linewidth]{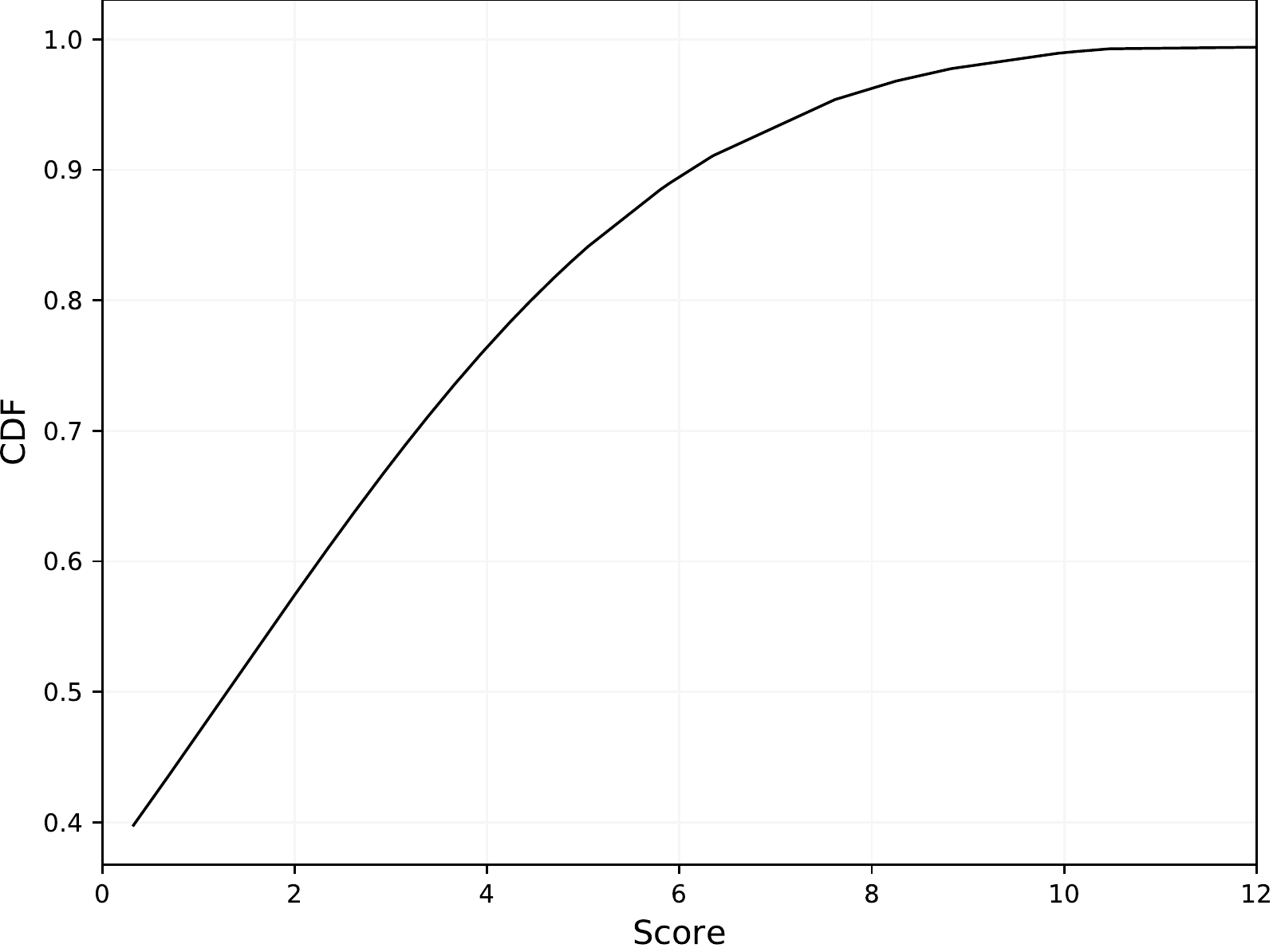}
        \caption{for novel words}
        \label{novel_cdf}
    \end{subfigure}
    \begin{subfigure}[b]{0.4\textwidth}
        \includegraphics[width=1\linewidth]{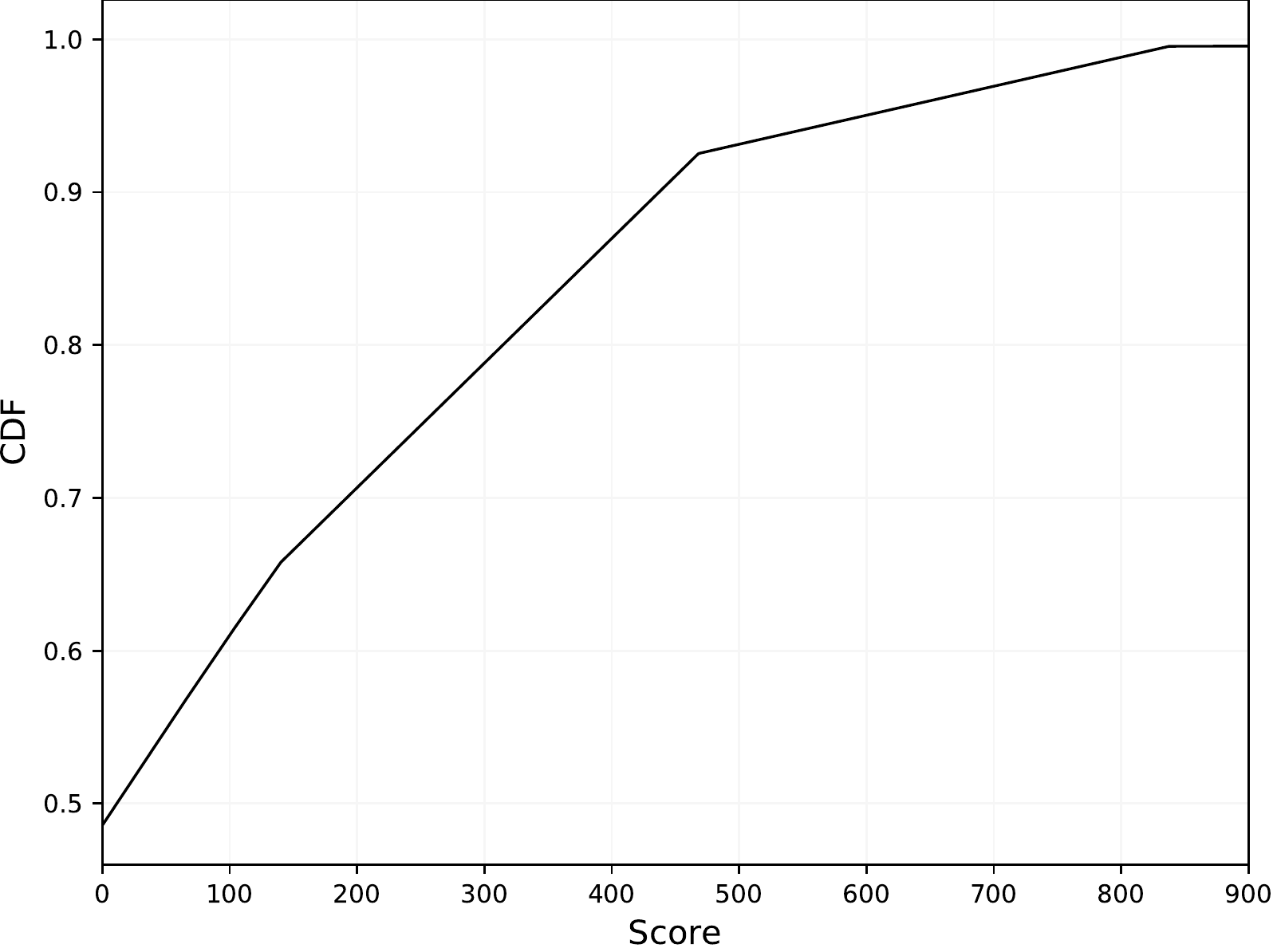}
        \caption{for shared words}
        \label{common_cdf}
    \end{subfigure}
    \caption{Cumulative frequency distribution (CFD) graph for word scores in different categories of BBC Sports using TM.}
    \label{bbc_cdf_graph_tm}
\end{figure}

\begin{figure}[h]
    \centering
    \begin{subfigure}[b]{0.4\textwidth}
        \includegraphics[width=1\linewidth]{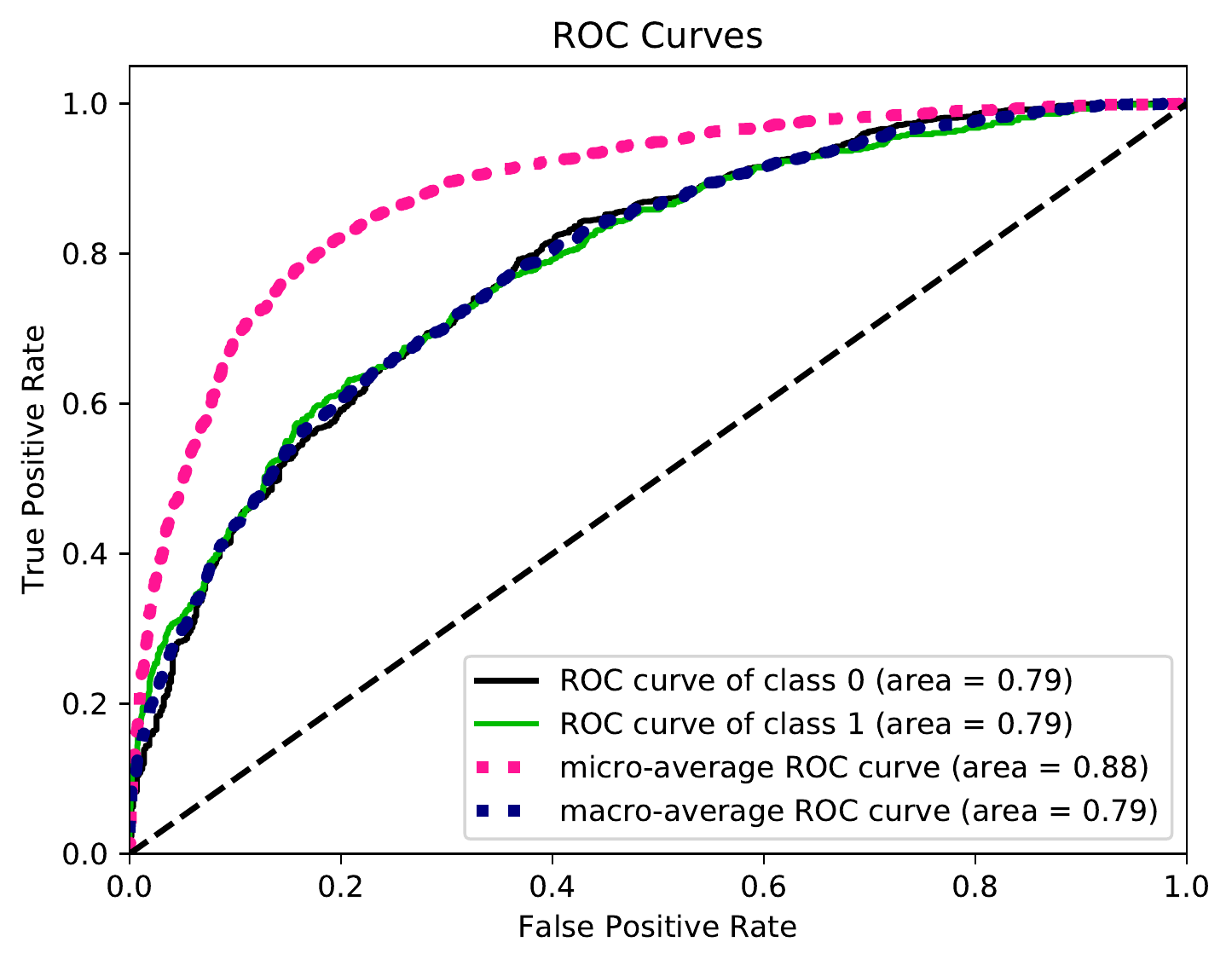}
        \caption{ROC curve}
        \label{bbc_roc_curve}
    \end{subfigure}
    \begin{subfigure}[b]{0.4\textwidth}
        \includegraphics[width=1\linewidth]{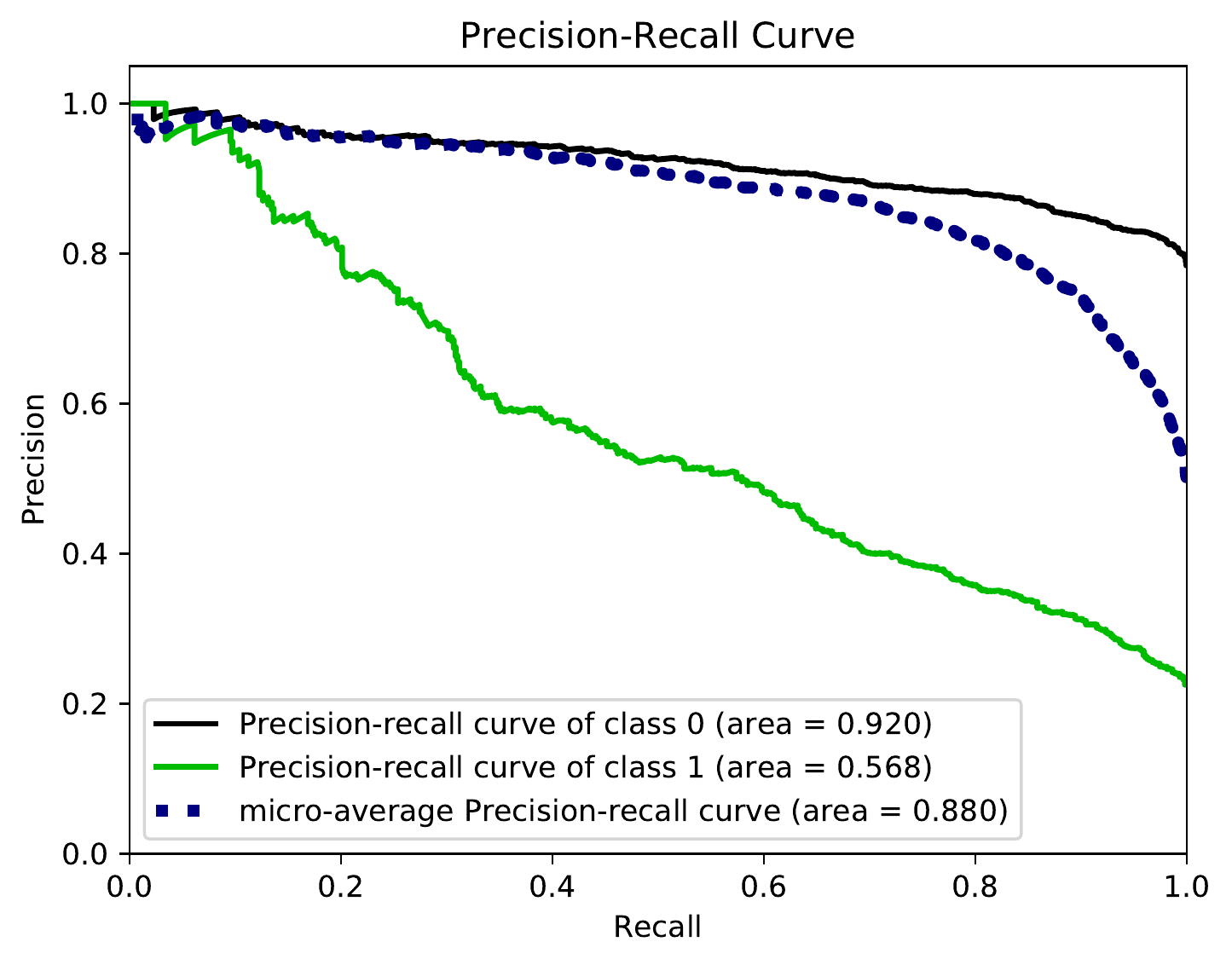}
        \caption{precision-recall graph}
        \label{bbc_precision_recall}
    \end{subfigure}
    \caption{ROC curve and precision-recall of known/novel class classification of BBC Sports using word scores obtained from TM.}
    \label{bbc_roc_logistic}
\end{figure}

\begin{figure}[h]
    \centering
    \begin{subfigure}[b]{0.4\textwidth}
        \includegraphics[width=1\linewidth]{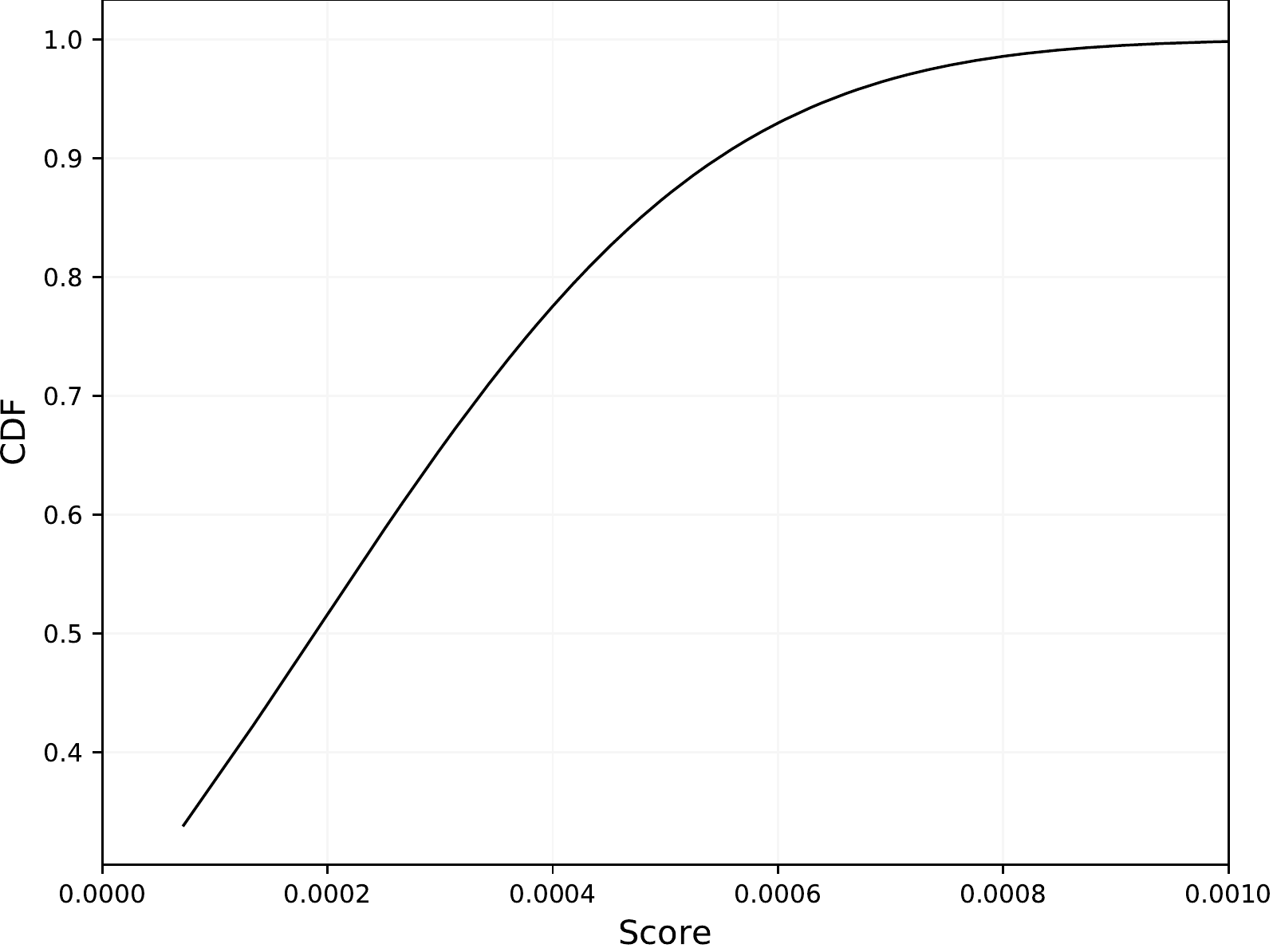}
        \caption{For known words.}
        \label{bbc_known_tfidf}
    \end{subfigure}
    \begin{subfigure}[b]{0.4\textwidth}
        \includegraphics[width=1\linewidth]{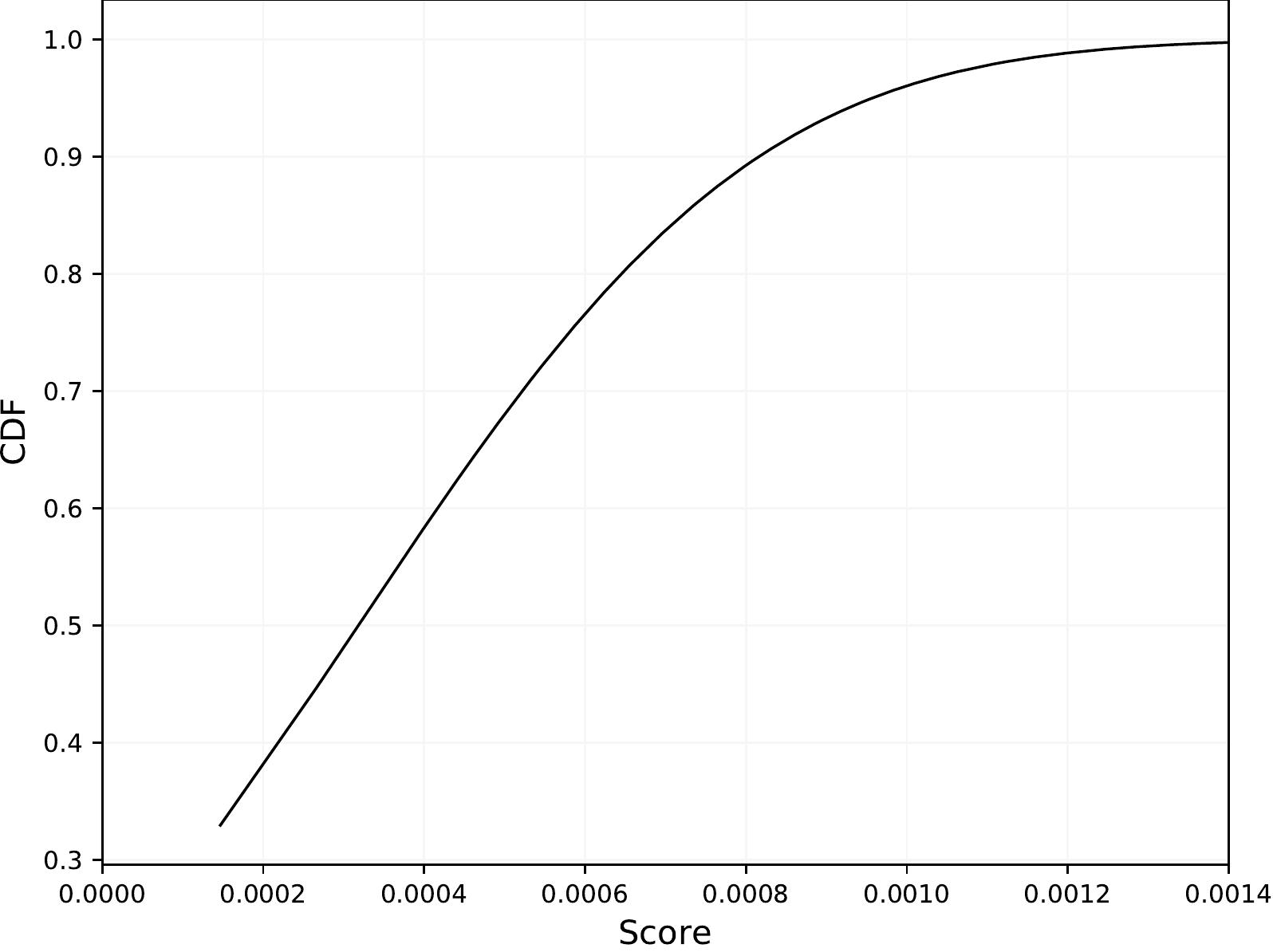}
        \caption{For novel words}
        \label{bbc_novel_tfidf}
    \end{subfigure}
    \begin{subfigure}[b]{0.4\textwidth}
        \includegraphics[width=1\linewidth]{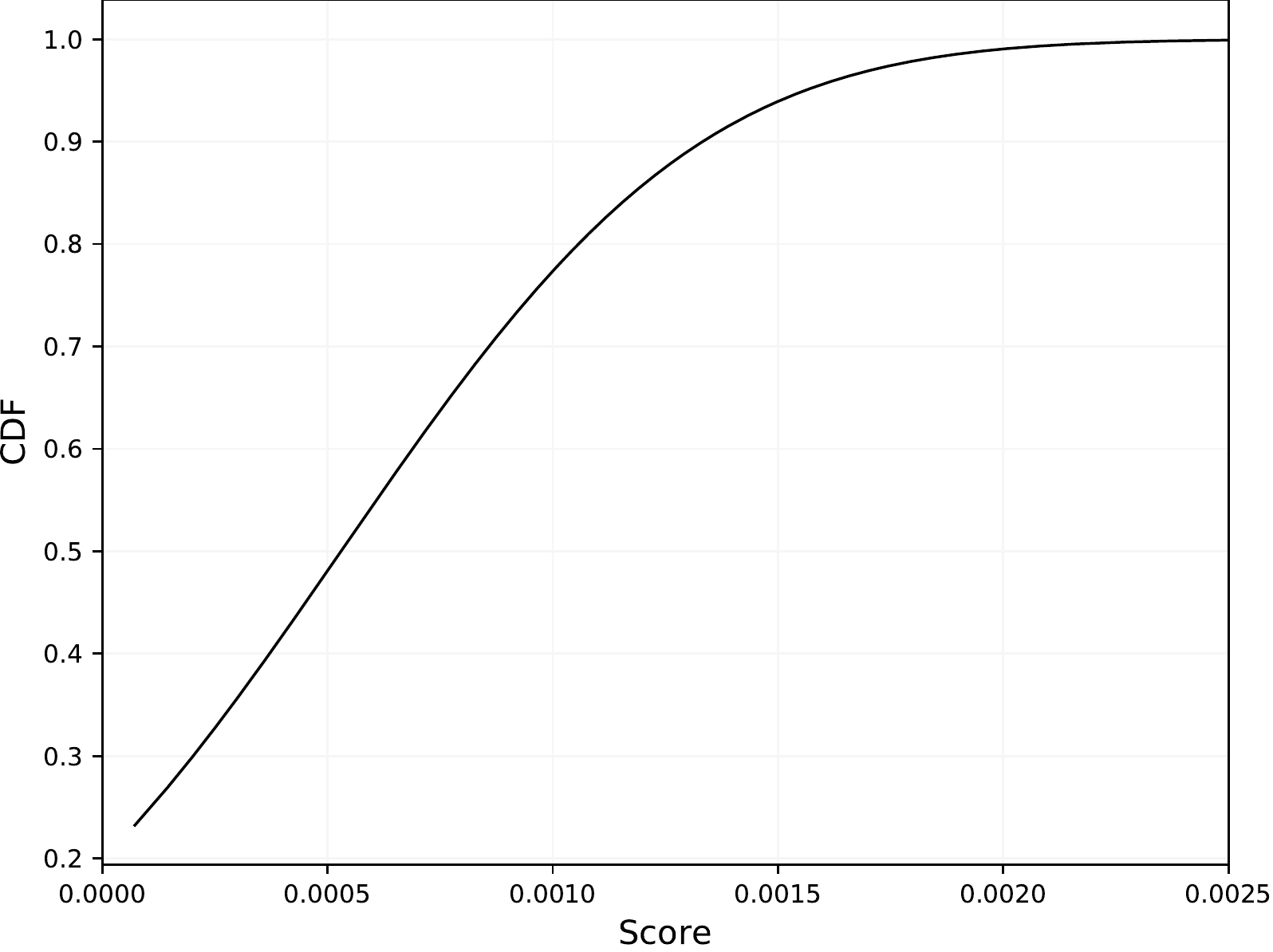}
        \caption{for shared words using known scores}
        \label{bbc_similar_known_tfidf}
    \end{subfigure}
    \begin{subfigure}[b]{0.4\textwidth}
        \includegraphics[width=1\linewidth]{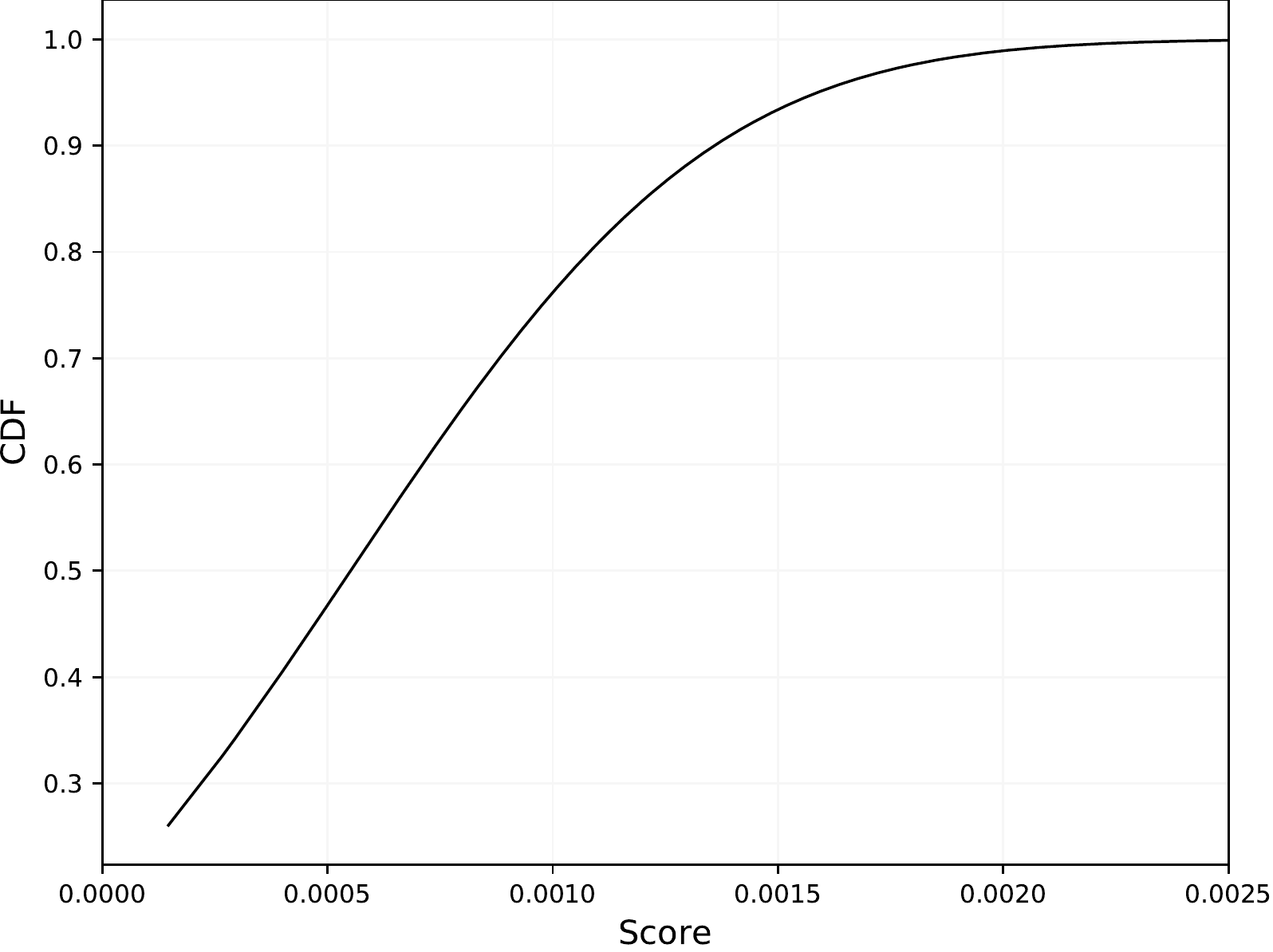}  
        \caption{for shared words using novel scores}
        \label{bbc_similar_novel_tfidf}
    \end{subfigure}
    \caption{Cumulative frequency distribution (CFD) graph for TF-IDF scores in different categories of BBC Sports.}
    \label{bbc_tfidf}
\end{figure}

\begin{figure}[h]
    \centering
    \begin{subfigure}[b]{0.4\textwidth}
        \includegraphics[width=1\linewidth]{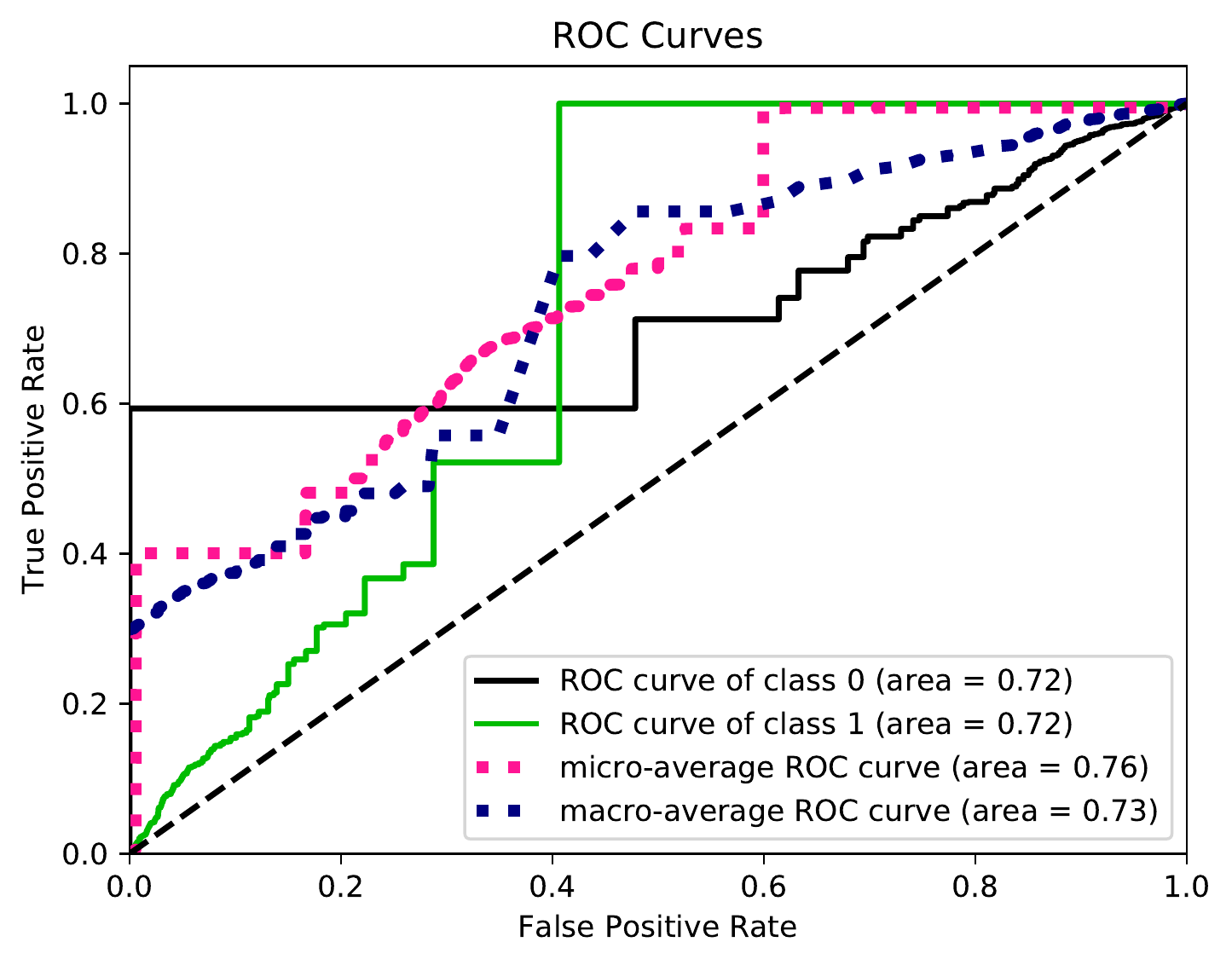}
        \caption{ROC curve}
        \label{bbc_roc_tfidf}
    \end{subfigure}
    \begin{subfigure}[b]{0.4\textwidth}
        \includegraphics[width=1\linewidth]{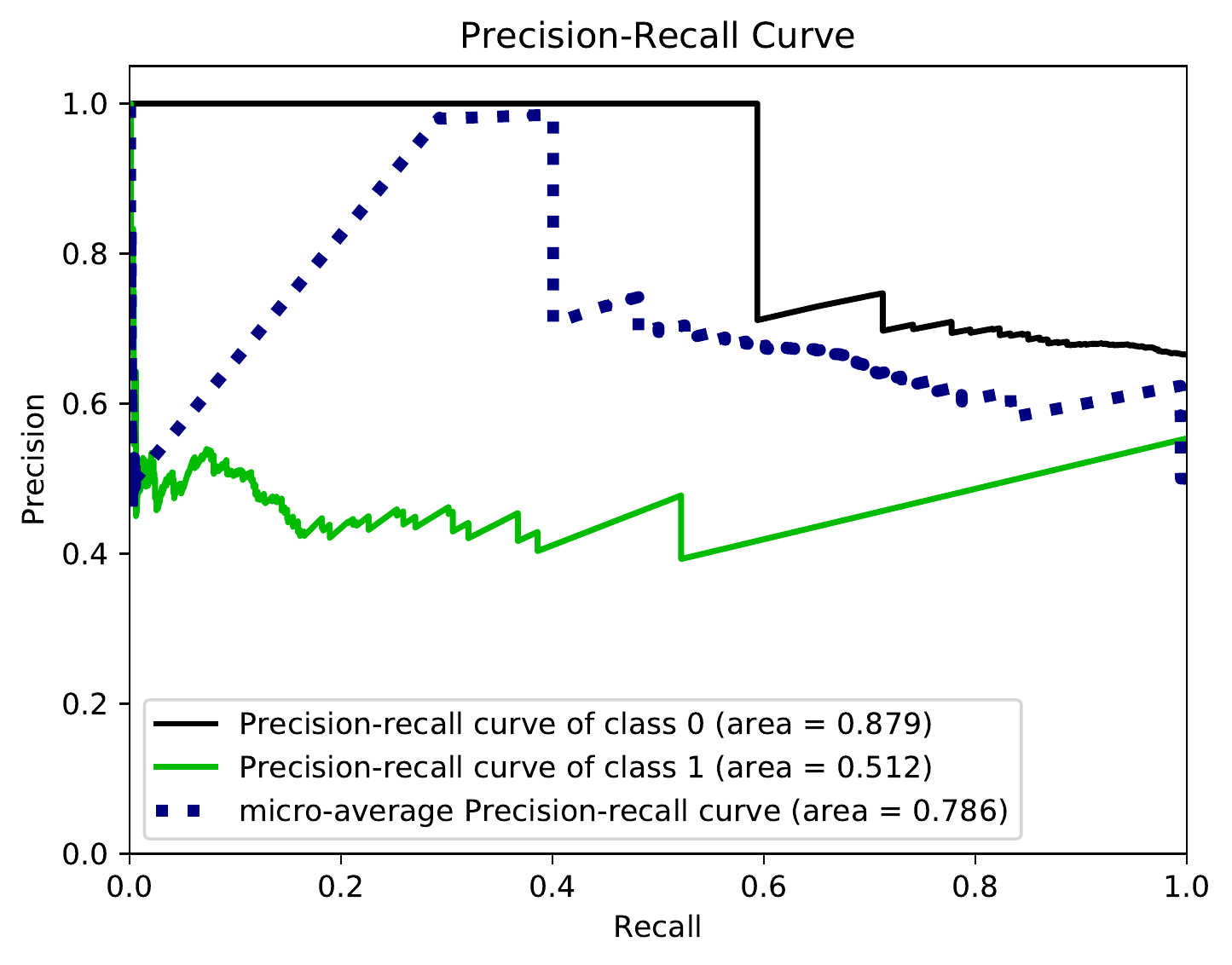}
        \caption{precision-recall graph}
        \label{bbc_precision_recall_tfidf}
    \end{subfigure}
    \caption{ROC curve and precision-recall of known/novel class classification of BBC Sports using TF-IDF scores.}
    \label{bbc_roc_tfidf_total}
\end{figure}

To gain further insight into the properties of the novelty score, we plot the CFD for the scores of the novel, known and shared words in Figure \ref{bbc_cdf_graph_tm}. We further compare these CDFs with the corresponding ones obtained using TF-IDF in Figure \ref{bbc_tfidf}. As seen, the plots confirm that our approach produces more distinctive novelty scores than TF-IDF. The novel words typically produce high scores, while the known words produce low scores. In particular, as shown in Figure \ref{known_cdf}, $85\%$ of the known words output scores lower than $1.0$. In Figure~\ref{novel_cdf}, on the other hand, we see that only about $45\%$ of the words unique for the novel class have scores below $1$. The majority of the uniquely novel words produce scores greater than $1$.

Finally, in Figure~\ref{common_cdf}, we plot the scores for words that are shared between the known and novel classes. As seen, the words that are shared produce both high and low scores. To cast further light on this observation, we investigate the words that are shared further in Table~\ref{overlapped_bbc}. We see that the words that are captured frequently by novel clauses have high scores, while the words that are frequent in known clauses have low scores. Further, common words (e.g. stopwords), also have low scores.  For example, the word ``Rugby'', which is highly characteristic for class \emph{Novel}, is repeated only $5$ times in the clauses representing class \emph{Known}. For the clauses that represent class \emph{Novel}, on the other hand, it is repeated $215$ times. In other words, the shared words constitutes words that are either characteristic for class \emph{Known} or for class \emph{Novel}. This finding also suggests that the scores can be calculated accurately even if the words are present in both categories.
\begin{table}
  \caption{Overall word statistics for 20 Newsgroups dataset}
  \label{newsgroup_table}
  \begin{tabular}{cccc}
    \toprule
    Category & Total word count & Average score & Standard deviation \\
    \midrule
    Known words & 23133 & 0.99 & 0.21 \\
    Novel words & 6921 & 1.20 & 1.04\\
    Shared words & 5786 & 3.04 & 131.62\\
  \bottomrule
\end{tabular}
\end{table}

\begin{table}
  \caption{Composition of shared words in 20 Newsgroups dataset}
  \label{overlapped_20newsgroup}
  \begin{tabular}{cccc}
    \toprule
    Composition & Total word count & Average score & Standard deviation \\
    \midrule
    Known words & 9 & 0.14 & 0.074 \\
    Novel words & 33 & 640.75 & 2378.87\\
    Common words & 5697 & 1.11 & 0.58\\
  \bottomrule
\end{tabular}
\end{table}

\begin{figure}[h]
    \centering
    \begin{subfigure}[b]{0.4\textwidth}
        \includegraphics[width=1\linewidth]{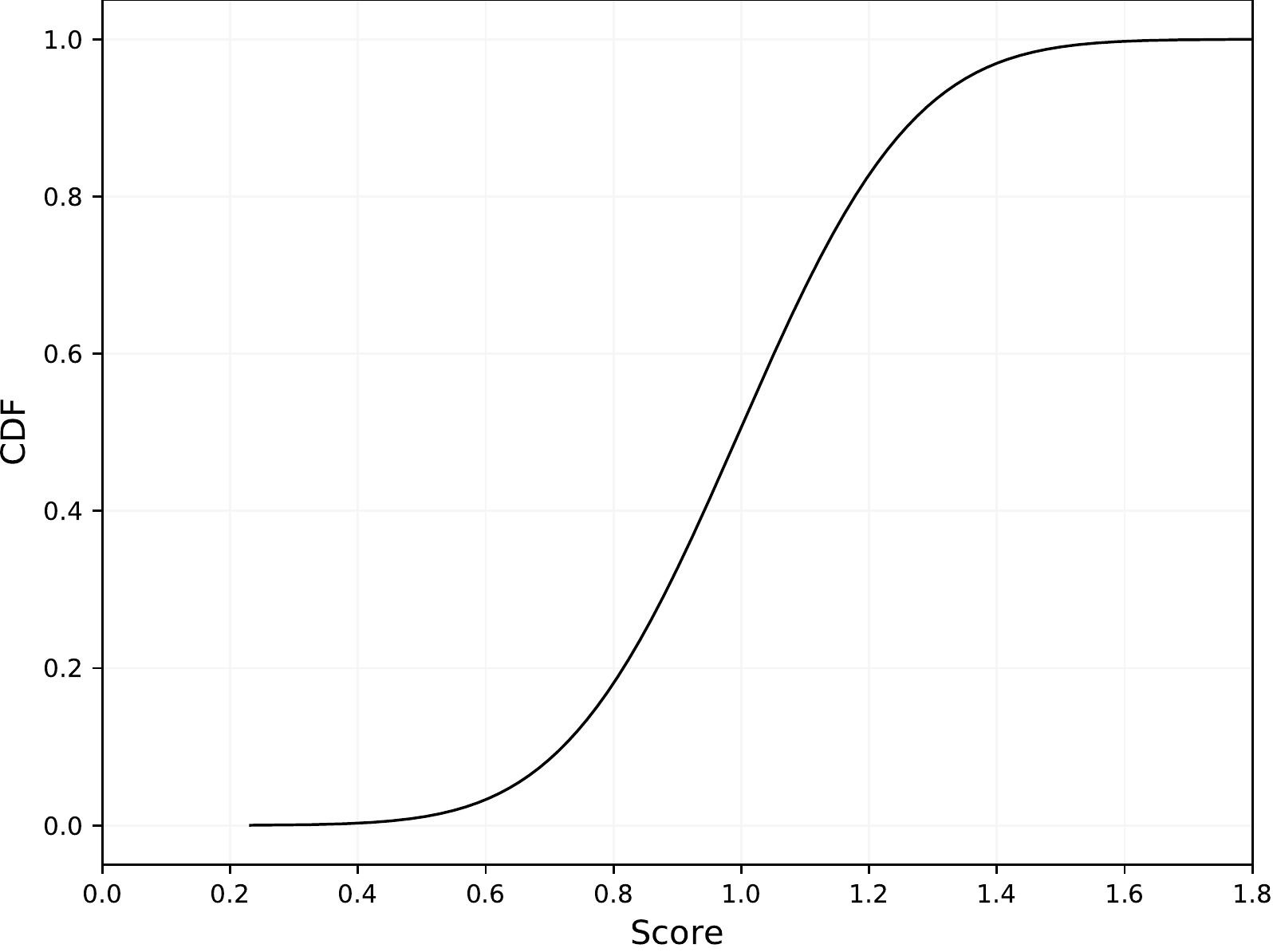}
        \caption{for known words}
        \label{newsgroup_known_cdf}
    \end{subfigure}
    \begin{subfigure}[b]{0.4\textwidth}
        \includegraphics[width=1\linewidth]{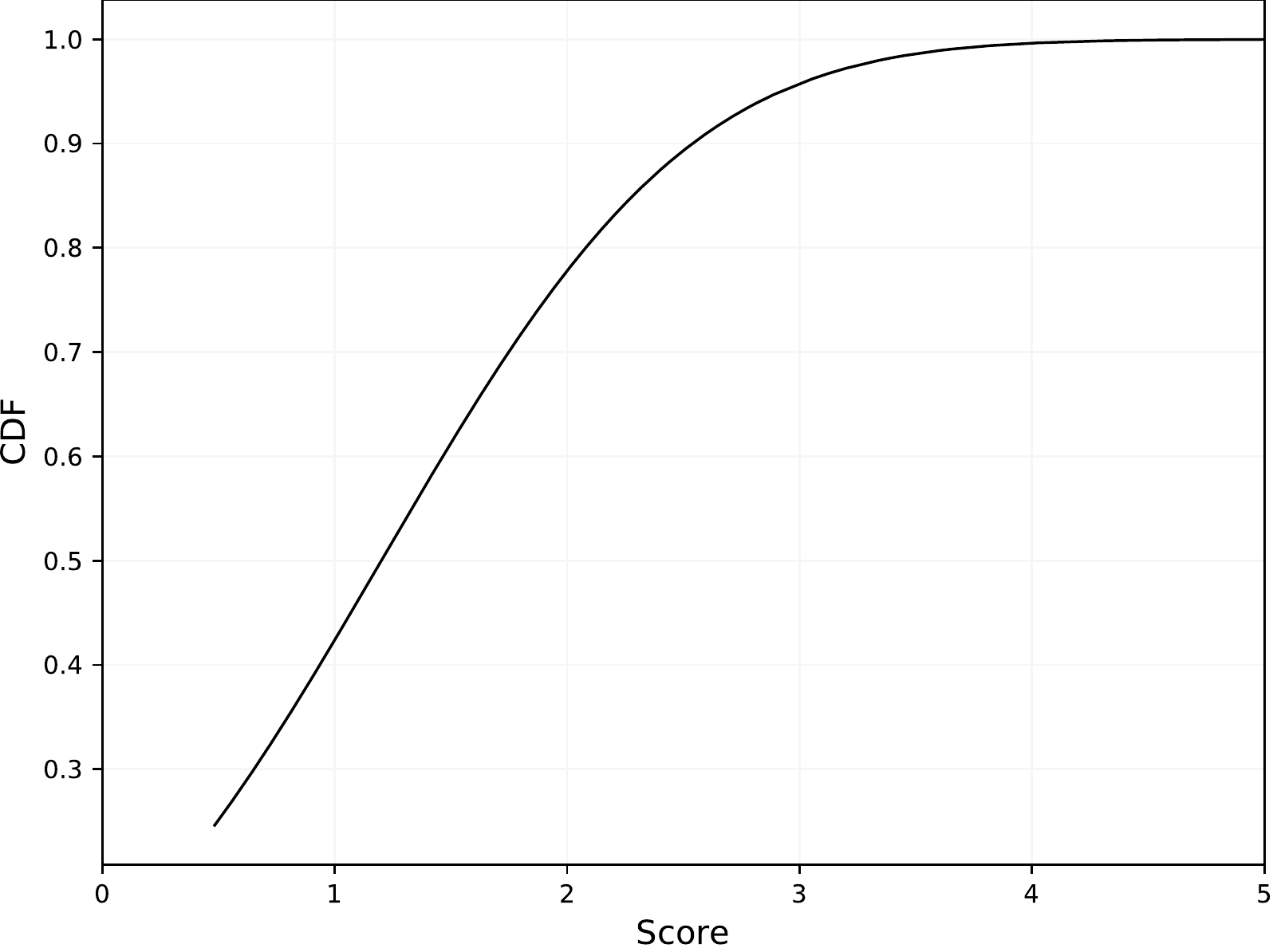}
        \caption{for novel words}
        \label{newsgroup_novel_cdf}
    \end{subfigure}
    \begin{subfigure}[b]{0.4\textwidth}
        \includegraphics[width=1\linewidth]{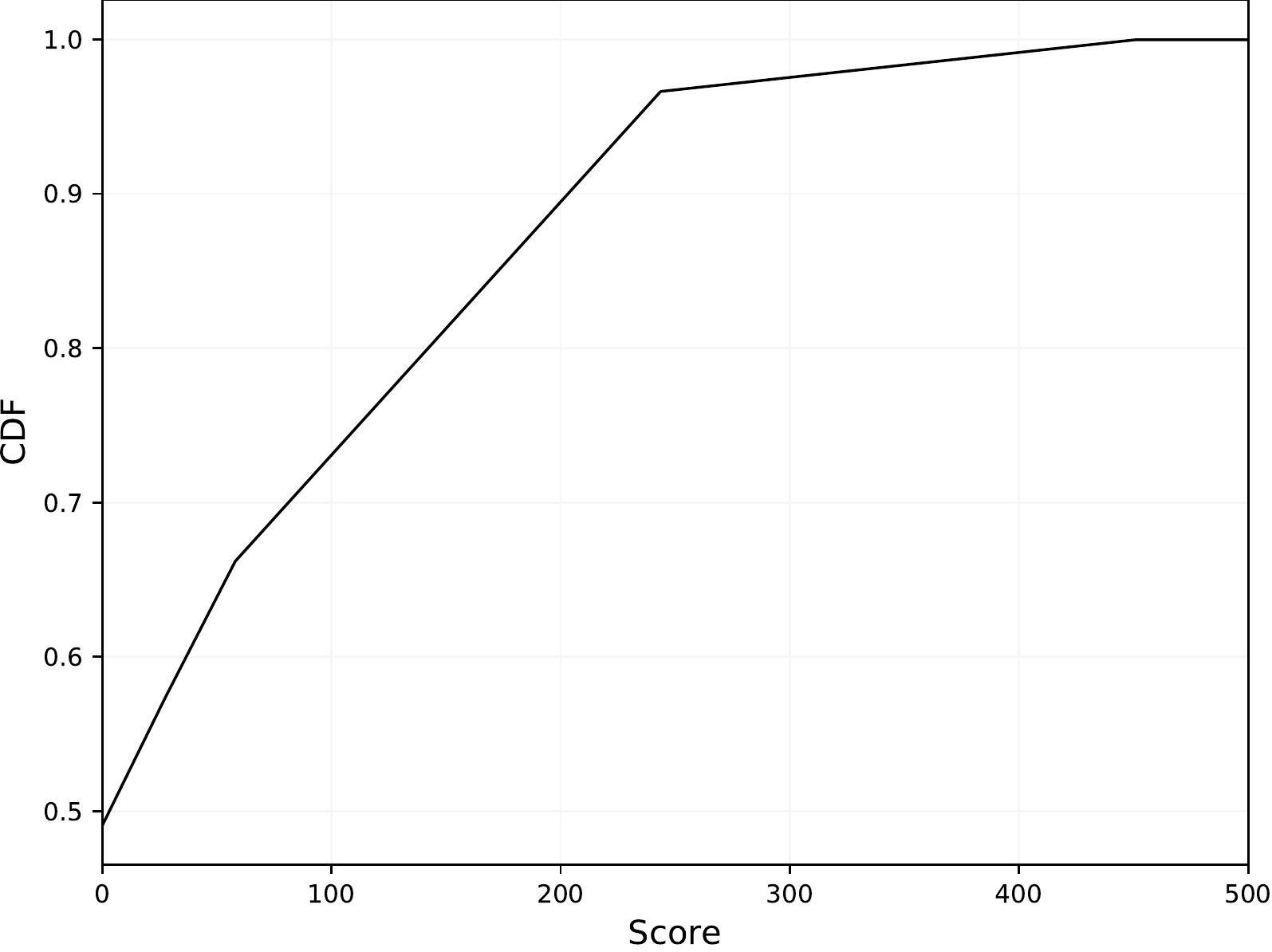}
        \caption{for common words}
        \label{newsgroup_common_cdf}
    \end{subfigure}
    \caption{Cumulative frequency distribution (CFD) graph for word scores in different categories of 20 Newsgroups using TM.}
    \label{newsgroup_cdf_graph}
\end{figure}

\begin{figure}[h]
    \centering
    \begin{subfigure}[b]{0.4\textwidth}
        \includegraphics[width=1\linewidth]{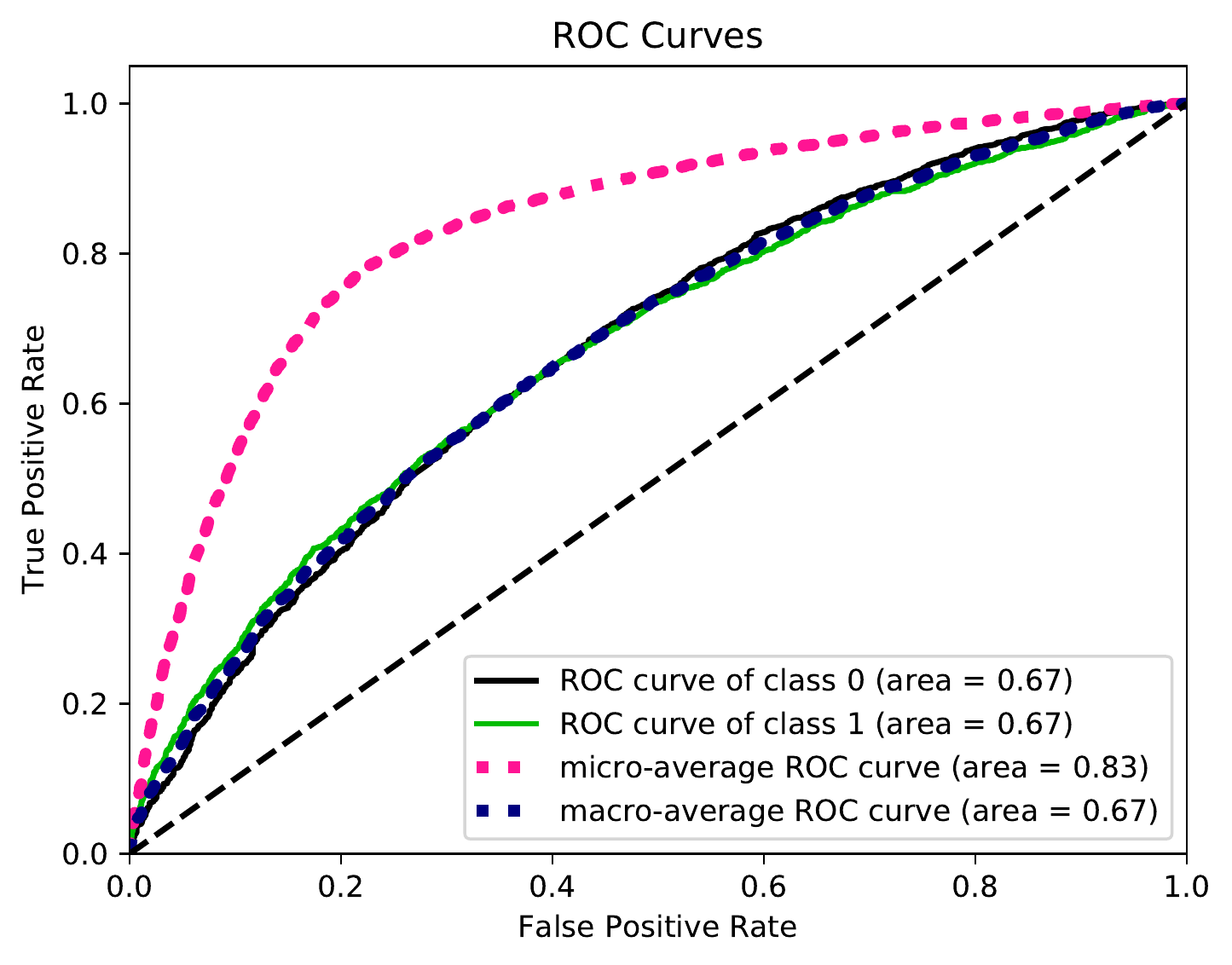}
        \caption{ROC curve}
        \label{newsgroup_roc_curve}
    \end{subfigure}
    \begin{subfigure}[b]{0.4\textwidth}
        \includegraphics[width=1\linewidth]{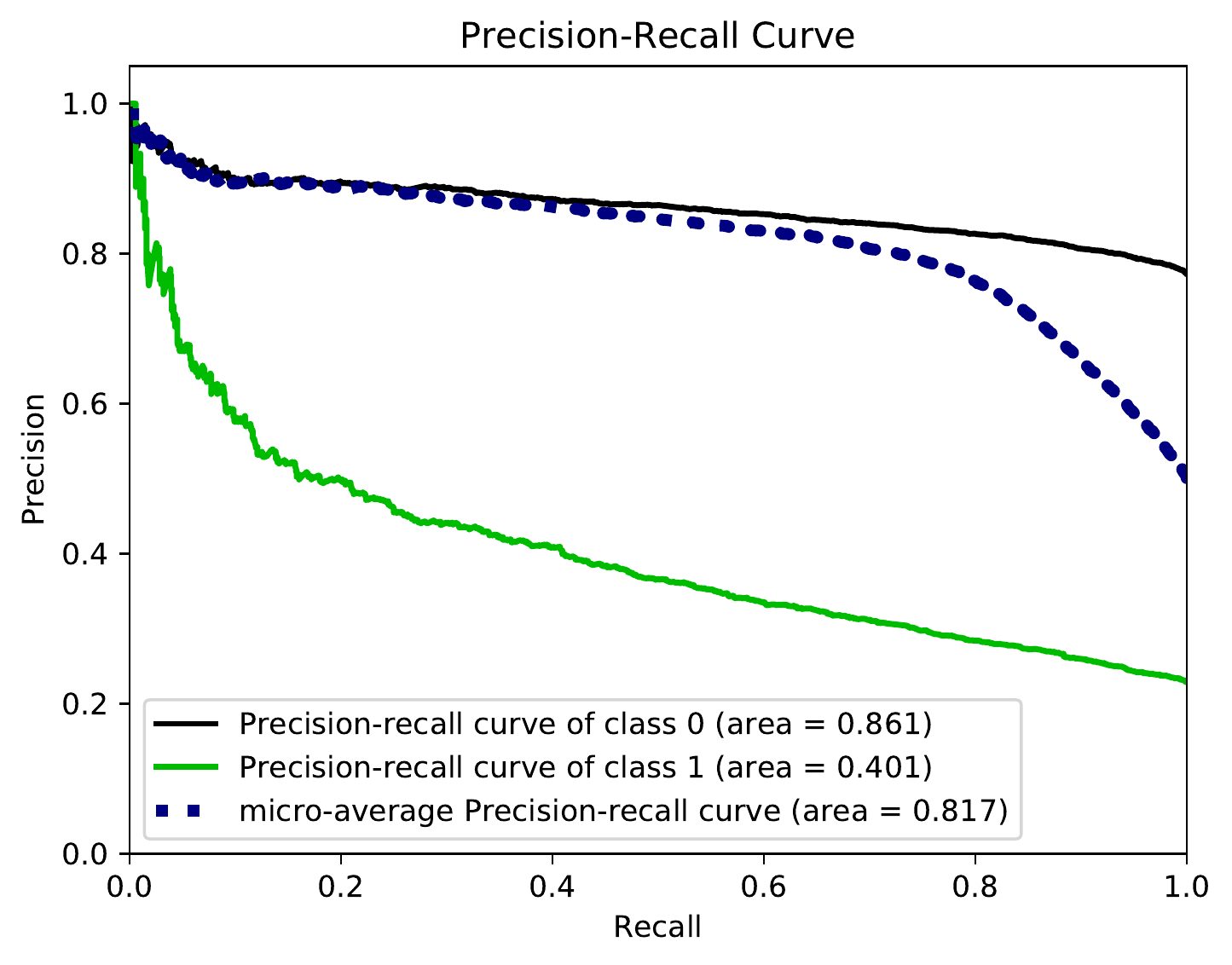}
        \caption{precision-recall graph}
        \label{newsgroup_precision_recall}
    \end{subfigure}
    \caption{ROC curve and precision-recall of known/novel class classification of 20 Newsgroups using word scores obtained from TM.}
    \label{newsgroup_roc_logistic}
\end{figure}

\begin{figure}[h]
    \centering
    \begin{subfigure}[b]{0.4\textwidth}
        \includegraphics[width=1\linewidth]{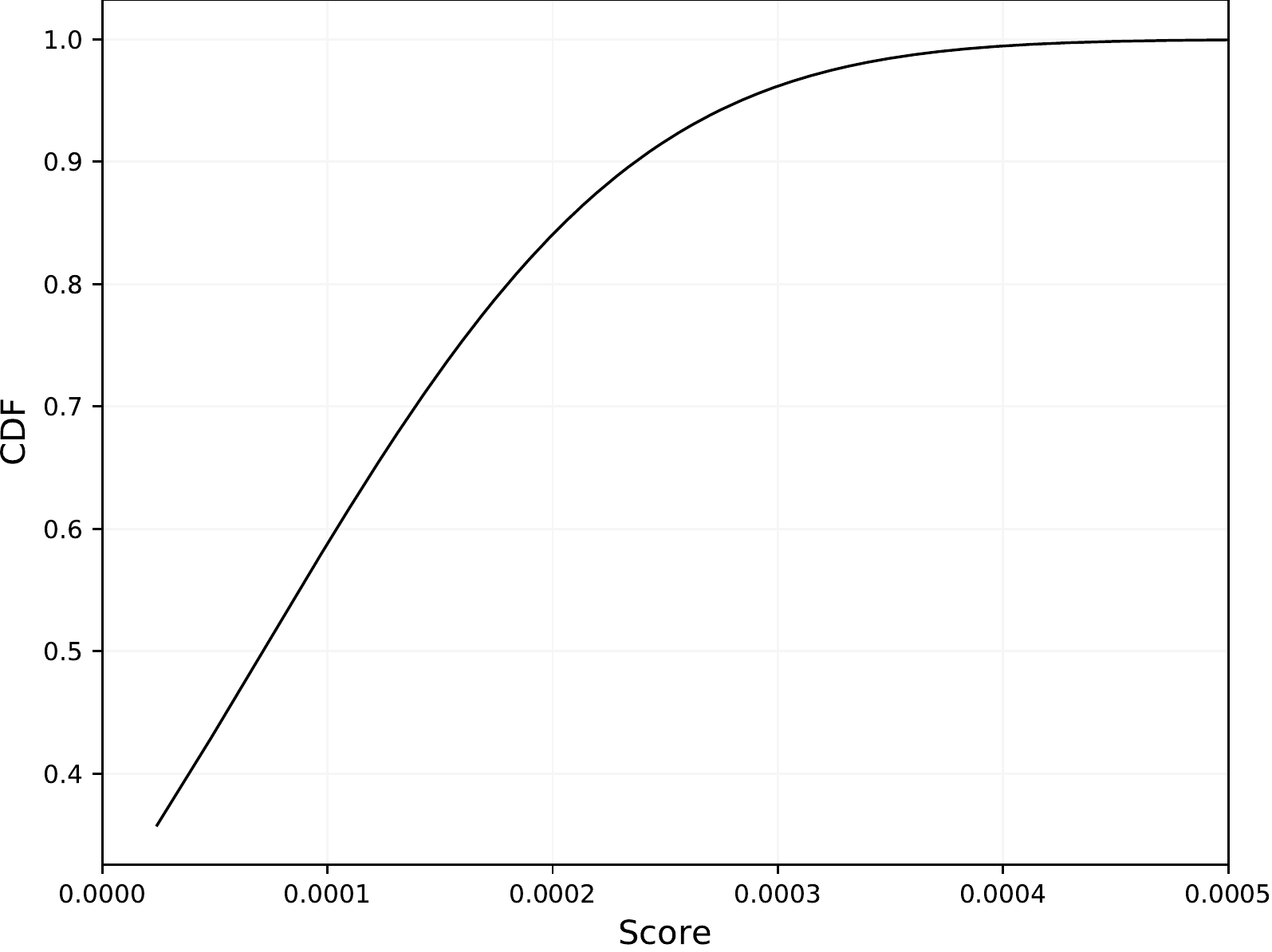}
        \caption{For known words.}
        \label{newsgroup_known_tfidf}
    \end{subfigure}
    \begin{subfigure}[b]{0.4\textwidth}
        \includegraphics[width=1\linewidth]{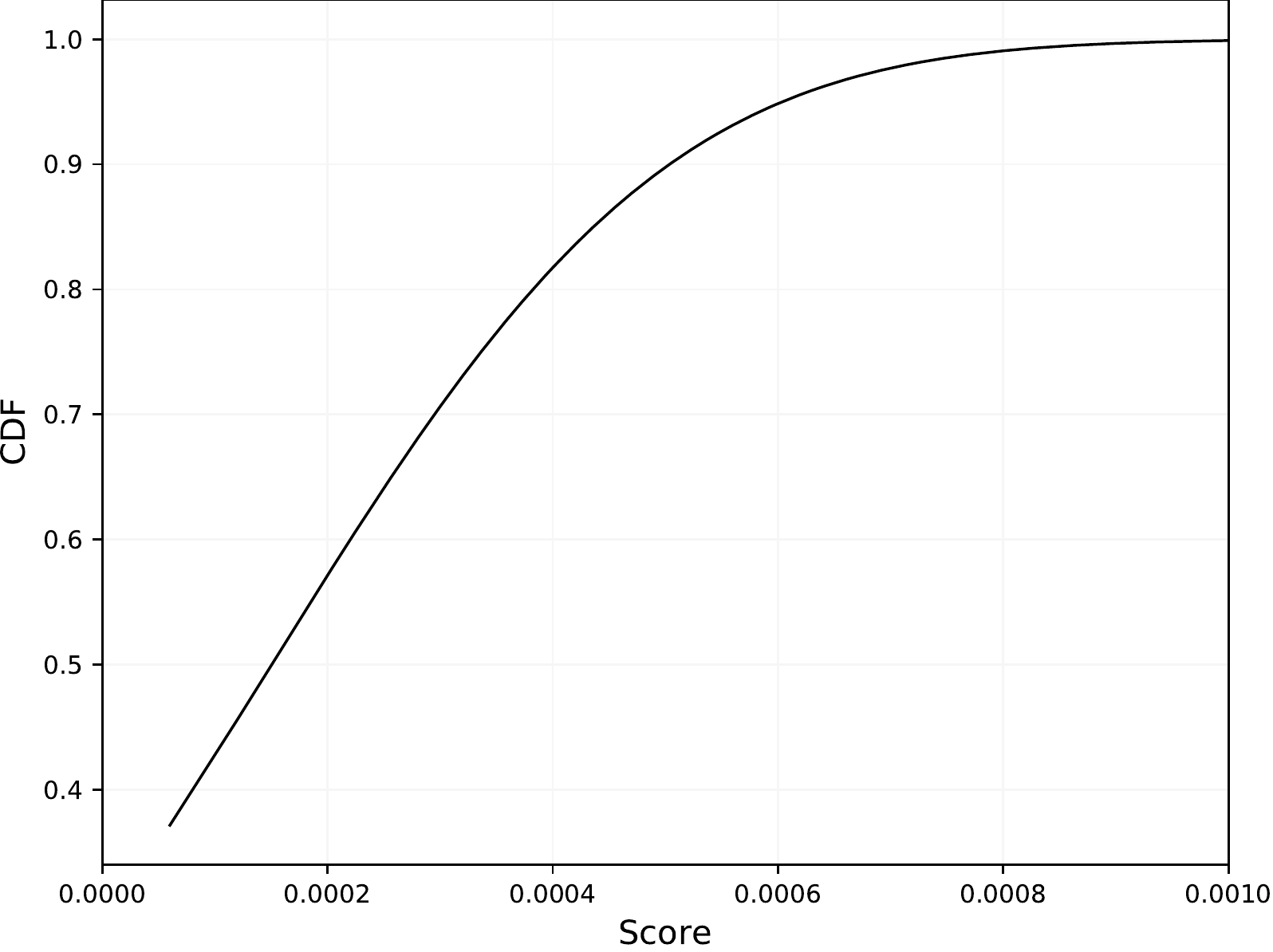}
        \caption{For novel words}
        \label{newsgroup_novel_tfidf}
    \end{subfigure}
    \begin{subfigure}[b]{0.4\textwidth}
        \includegraphics[width=1\linewidth]{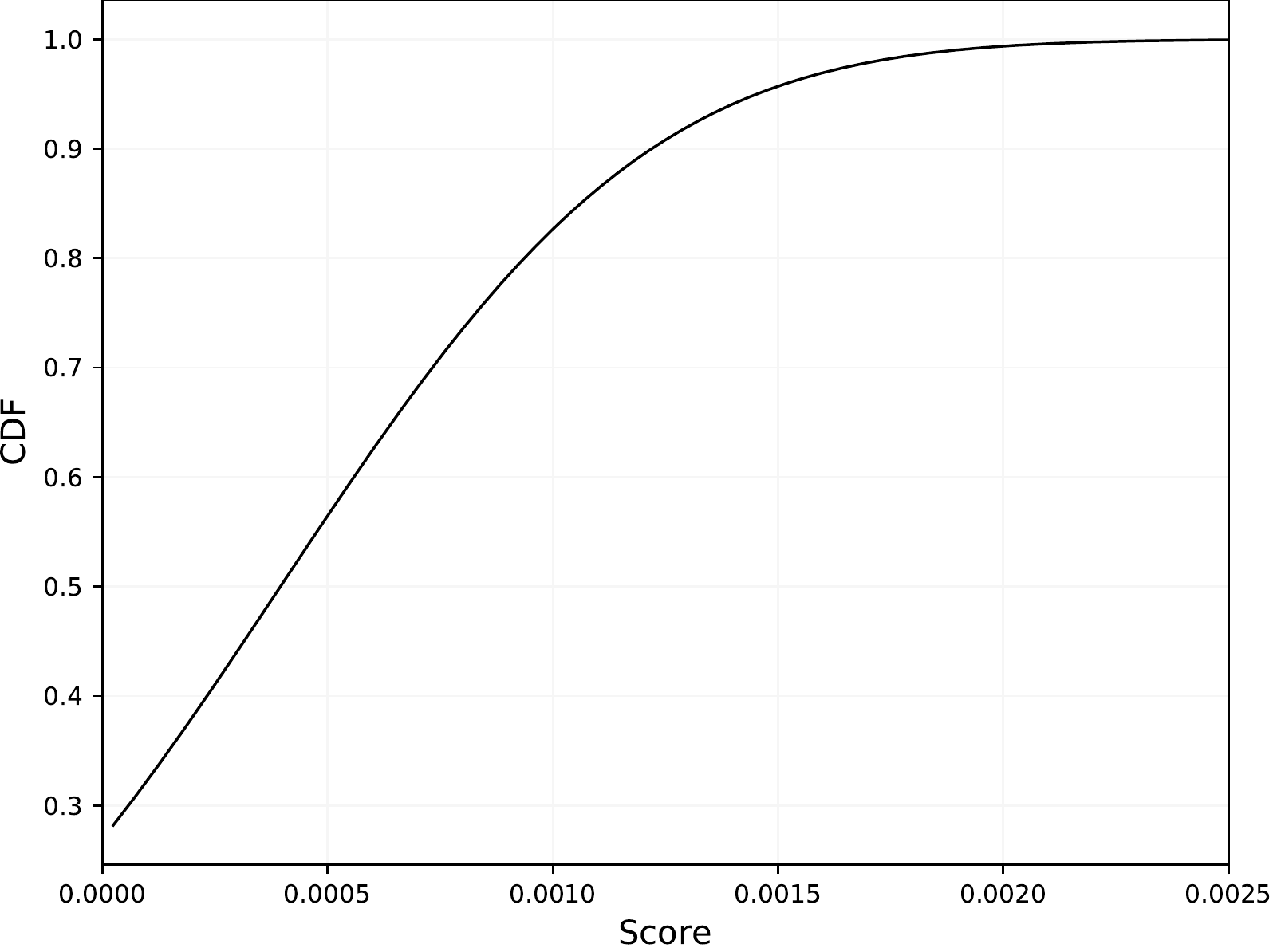}
        \caption{for shared words using known scores}
        \label{newsgroup_similar_known_tfidf}
    \end{subfigure}
    \begin{subfigure}[b]{0.4\textwidth}
        \includegraphics[width=1\linewidth]{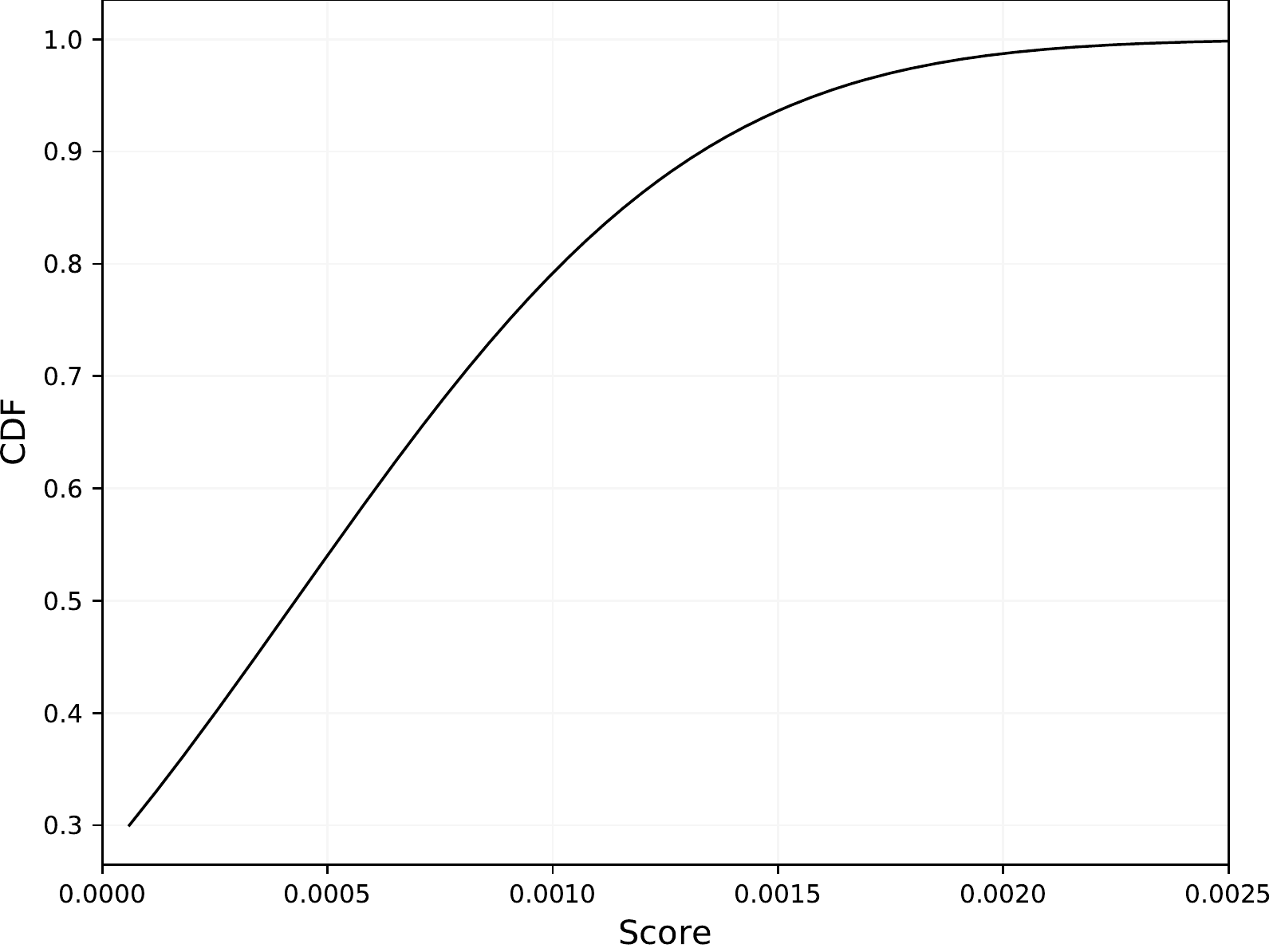}
        \caption{for shared words using novel scores}
        \label{newsgroup_similar_novel_tfidf}
    \end{subfigure}
    \caption{Cumulative frequency distribution (CFD) graph for TFIDF scores in different categories of 20 Newsgroups.}
    \label{newsgroup_tfidf}
\end{figure}

\begin{figure}[h]
    \centering
    \begin{subfigure}[b]{0.4\textwidth}
        \includegraphics[width=1\linewidth]{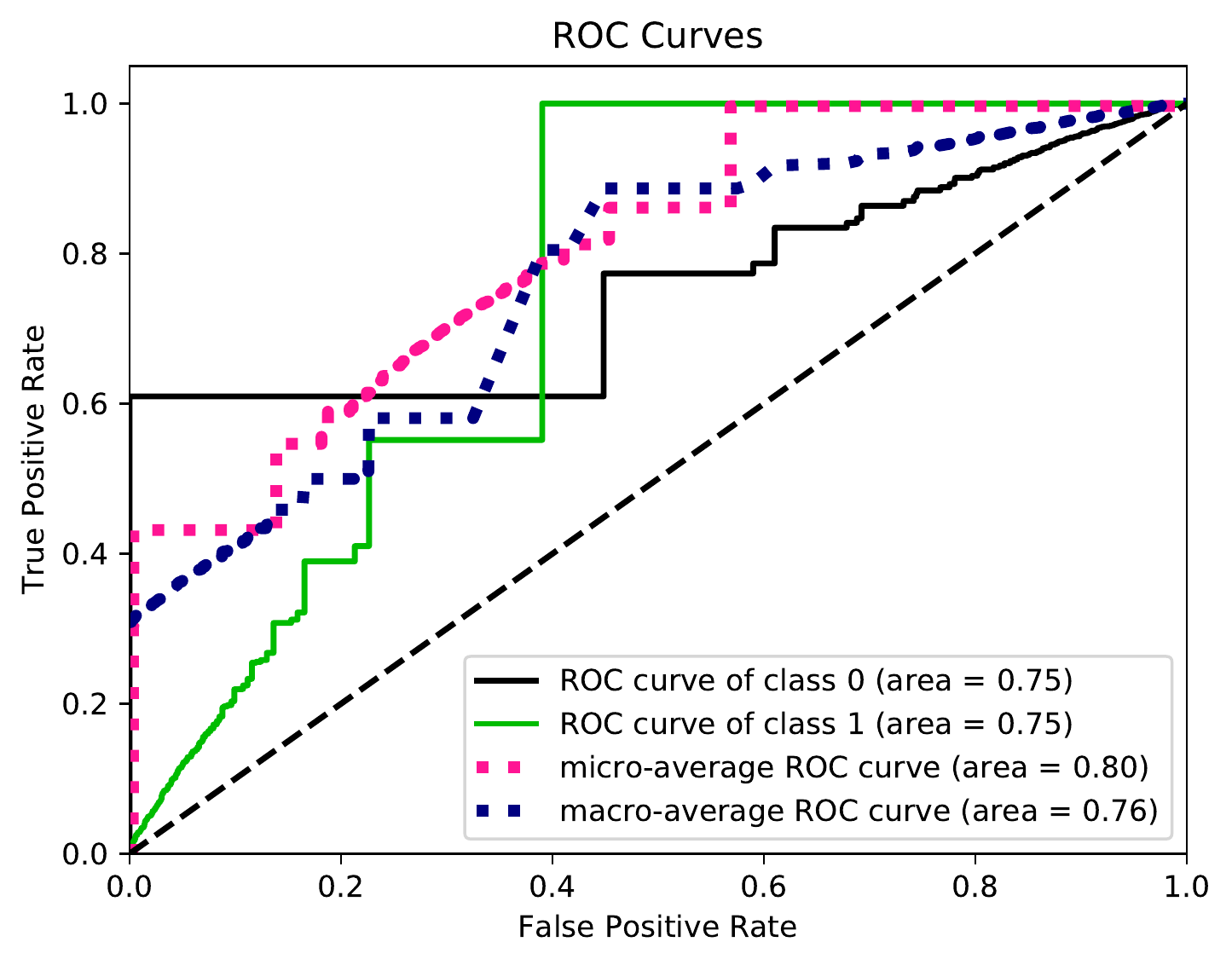}
        \caption{ROC curve}
        \label{newsgroup_roc_curve_tfidf}
    \end{subfigure}
    \begin{subfigure}[b]{0.4\textwidth}
        \includegraphics[width=1\linewidth]{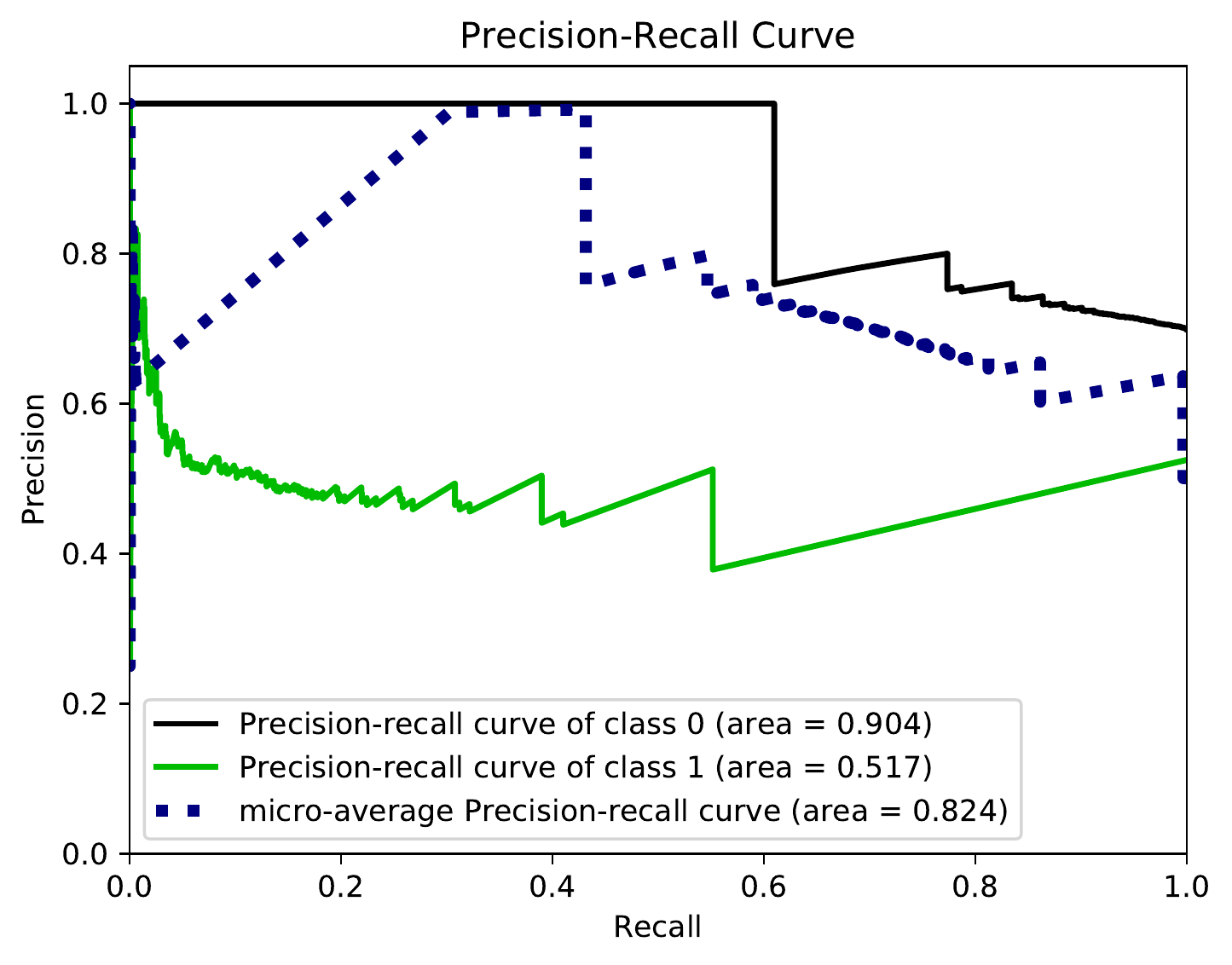}
        \caption{precision-recall graph}
        \label{newsgroup_precision_recall_tfidf}
    \end{subfigure}
    \caption{ROC curve and precision-recall of known/novel class classification of 20 Newsgroups using TFIDF scores.}
    \label{newsgroup_roc_logistic_tfidf}
\end{figure}

We now investigate the degree of discrimination power our novelty scoring provides, and thus uniquely characterizes novelty at the word-level. To this end, we employ logistic regression for classifying novel text based on the word scores obtained from our method. The ROC and precision-recall curves of the experiment are depicted in Figure \ref{bbc_roc_logistic} for our novelty scoring mechanism. Figure \ref{bbc_roc_tfidf_total} contains corresponding curves when TF-IDF scores are used instead. We see that the classification performance for our novelty scores are significantly better than what is obtained with TF-IDF.


\subsection{20 Newsgroups dataset}

The 20 Newsgroups dataset contains a total of $18~828$ documents partitioned equally into $20$ separate classes. In our experiments, we treat the two classes ``comp.graphics'' and ``talk.politics.guns'' as \emph{Known} topics, and  then use the class ``rec.sport.baseball'' to represent a \emph{Novel} topic. Again, we train a TM to produce our clause-based novelty scores. The overall statistics of the resulting word scores are shown in Tables \ref{newsgroup_table} and \ref{overlapped_20newsgroup}, where we observe similar behaviour as for the BBC Sports dataset.

The CFD plot in Figure \ref{newsgroup_cdf_graph} presents the score distribution among words per group (known, novel, shared). For known words, in Figure \ref{newsgroup_known_cdf}, we see that $90\%$ of the scores of the words are below around $1.3$. In Figure \ref{newsgroup_novel_cdf}, however, only $45\%$ of the novel word scores are below approx. $1.3$. From the plots, it is clear that most of the novel words have significantly higher scores than the known words. Note that some of the low scores of some novel words are due to the common words (e.g. stop words) present in the novel bag-of-words. Since the common words, as such, do not signify novelty, the TM clauses do not frequently capture them. Hence, they provide relatively low scores despite only appearing among the novel documents. Finally, we again observe that the shared words have been used by the clauses for discrimination (cf. Table \ref{overlapped_20newsgroup}), hence provides a mix of low and high novelty scores, as shown in Figure~\ref{newsgroup_common_cdf}. Again, we observe similar behavior as for the BBC Sports dataset.

\subsection{Contextual scoring}
\begin{table}
    \centering
    \caption{Co-occurrence matrix showing the information gain between words in BBC Sports.}
    \label{bbc_neighbor}
    \renewcommand{\arraystretch}{1}
    \begin{tabular}{ll|l|l|l|l|l|}
        
\multicolumn{2}{c}{}&   \multicolumn{5}{c}{}\\
        \multicolumn{2}{c}{}&\multicolumn{5}{c}{{\rotatebox[origin=c]{0}{Manchester}
            } {\rotatebox[origin=c]{0}{Chelsea}
            } {\rotatebox[origin=c]{0}{Particular}
            } {\rotatebox[origin=c]{0}{Rugby}
            } {\rotatebox[origin=c]{0}{Flyhalf}
        }}\\
        \cline{3-7}
        \multirow{5}{*}{{\rotatebox[origin=c]{90}{}
        }} & 
        Manchester&14363.688&6.324&4.738&0.33&0.848  \\ \cline{3-7}
        &   Chelsea&6.324&19801.49&6.18&0.466&1.326 \\ \cline{3-7}
        &   Particular&4.738&6.18&30863.006&2.52&4.968 \\ \cline{3-7}
        &   Rugby&0.33&0.466&2.52&486.758&3.952 \\ \cline{3-7}
        &   Flyhalf&0.848&1.326&4.968&3.952&8888.888 \\ \cline{3-7}
    \end{tabular}
\end{table}

We also implement a context-based scoring approach to investigate how multiple words interact to capture novelty. As detailed in Section \ref{novelty_description}, we calculate our joint novelty score by measuring word co-occurrence in clauses. That is, we intend to capture how context can help uncover novelty when words have multiple meanings.  The context-based scoring is important because the context can change the word from being novel to known, such as the meaning of the word ``apple'' in ``apple fruit'' and ``apple phone''. For demonstration, we calculate our proposed context-based novelty score for five words (i.e., two known, two novel and one common word) in both datasets. For the BBC Sports dataset, the pairwise co-occurrence scores are shown in Table \ref{bbc_neighbor}. We see a high correspondence between words such as  ``Manchester'' and ``Chelsea'' from class \emph{Known}. Similarly, there is high correspondence between words such as ``Rugby'' and ``Flyhalf'' from class \emph{Novel}. The common word ``Particular'', on the other hand, shows similar correspondence with words from both of the classes. Similarly, for the 20 Newsgroups dataset, the co-occurrence scores for five words selected from the known, novel and common word types are shown in Table \ref{news_neighbor}. The words ``Guns'' and ``Weapon'' are from class \emph{Known} and manifest strong co-occurrence. Further, the words ``Baseball'' and ``Player'' from class \emph{Novel} correspond strongly as well. The common word ``Gather'', on the other hand, co-occurs within both of the classes. These examples demonstrate that the words that are most likely to appear in a same context have a high co-occurrence score. This can be explained by the fact that words that tend to appear together in a similar context are captured by many clauses.

\begin{table}
    \centering
    \caption{Co-occurrence matrix showing the information gain between most repeated words from known clauses in 20 Newsgroup.}
    \label{news_neighbor}
    \renewcommand{\arraystretch}{1}
    \begin{tabular}{ll|l|l|l|l|l|}
        
\multicolumn{2}{c}{}&   \multicolumn{5}{c}{}\\
        \multicolumn{2}{c}{}&\multicolumn{5}{c}{{\rotatebox[origin=c]{0}{Guns}
            } {\rotatebox[origin=c]{0}{Weapon}
            } {\rotatebox[origin=c]{0}{Gather}
            } {\rotatebox[origin=c]{0}{Baseball}
            } {\rotatebox[origin=c]{0}{Player}
        }}\\
        \cline{3-7}
        \multirow{5}{*}{{\rotatebox[origin=c]{90}{}
        }} & 
        Guns&12302.96&17.648&15.754&4.036&4.268  \\ \cline{3-7}
        &   Weapon&17.648&13888.888&12.108&4.66&5.102 \\ \cline{3-7}
        &   Gather&15.754&12.108&14610.272&11.854&15.408 \\ \cline{3-7}
        &   Baseball&4.036&4.66&11.854&4003.824&18.566 \\ \cline{3-7}
        &   Player&4.268&5.102&15.408&18.566&9255.402 \\ \cline{3-7}
    \end{tabular}
\end{table}

\section{Conclusion}
In this work, we propose a Tsetlin Machine (TM)-based solution for word-level novelty description. First, we employ the clauses from a trained TM to capture how the most significant words differentiate a group of novel documents apart from a group of known documents. Then, we calculate the score for each word based on the role it plays in the clauses. The analysis of our empirical results for BBC Sports and 20 Newsgroups demonstrate significantly better novelty discrimination power when compared to using TF-IDF. Our empirical results also show that we can capture word relations through a contextual scoring mechanism that measure co-occurrence within TM clauses. By capturing non-linear relationships among words, we can enhance the capability of measuring novelty at the word level. However, training a TM is computationally more expensive than calculating TF-IDF, in particular for large datasets with a large vocabulary. We will address computation speed in our future work, employing indexing mechanisms and exploiting feature space sparsity.



\bibliographystyle{ACM-Reference-Format}
\bibliography{sample-base}

\end{document}